\documentclass[final]{cvpr}

\usepackage{times}
\usepackage{epsfig}
\usepackage{graphicx}
\usepackage{amsmath,amsfonts,amsthm,bm}
\usepackage{amssymb}
\usepackage{comment}
\usepackage{float}
\usepackage[dvipsnames]{xcolor}
\usepackage{color, colortbl}
\usepackage{arydshln}
\usepackage{placeins}
\usepackage{array,booktabs,ragged2e}

\newcolumntype{P}[1]{>{\centering\arraybackslash}p{#1}}
\usepackage[pagebackref=true,breaklinks=true,colorlinks,bookmarks=false]{hyperref}

\def\cvprPaperID{1180} 
\def\confYear{CVPR 2021}
\newcommand{\methodHz}{35}
\begin{document}
\title{SceneGraphFusion: Incremental 3D Scene Graph Prediction \\from RGB-D Sequences} 
\author{Shun-Cheng Wu\textsuperscript{1}
\quad
Johanna Wald\textsuperscript{1}
\quad
Keisuke Tateno\textsuperscript{2}
\quad
Nassir Navab\textsuperscript{1}
\quad
Federico Tombari\textsuperscript{1,2}\\
\textsuperscript{1}Technische Universit\"{a}t M\"{u}nchen \hspace{1.5cm} \textsuperscript{2}Google\\[0.1cm]
{\small \href{https://shunchengwu.github.io/SceneGraphFusion}{shunchengwu.github.io/SceneGraphFusion}}
}

\makeatletter
\g@addto@macro\@maketitle{
  \begin{figure}[H]
  \setlength{\linewidth}{\textwidth}
  \setlength{\hsize}{\textwidth}
  \centering
    \includegraphics[width=0.96\linewidth]{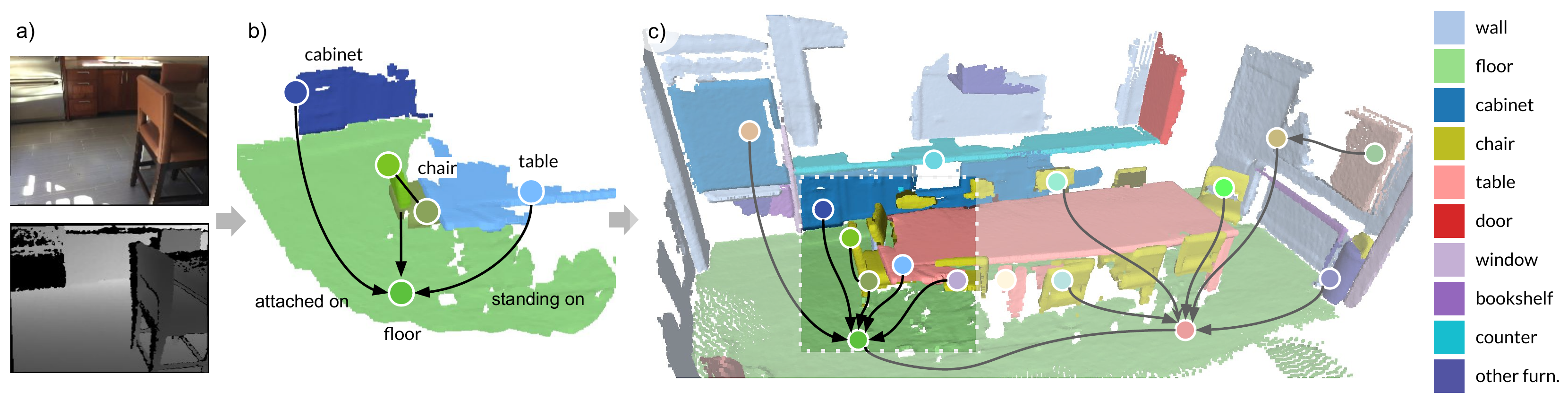}
	\setlength{\abovecaptionskip}{-8pt}
	\setlength{\belowcaptionskip}{8pt}
	\caption{We create a globally consistent 3D scene graph b) by fusing predictions of a graph neural network (GNN) from an incremental geometric segmentation created from an RGB-D sequence a). Our method merges nodes on the same object instance and naturally grows and improves over time when new segments and surfaces are discovered, see c). As a by-product, our method produces accurate panoptic segmentation of large-scale 3D scans. The nodes represent the different object segments.}
    \label{fig:teaser}
	\end{figure}
}
\makeatother
\maketitle
\thispagestyle{empty}
\begin{abstract}
   Scene graphs are a compact and explicit representation successfully used in a variety of 2D scene understanding tasks. This work proposes a method to incrementally build up semantic scene graphs from a 3D environment given a sequence of RGB-D frames. 
   To this end, we aggregate PointNet features from primitive scene components by means of a graph neural network. We also propose a novel attention mechanism well suited for partial and missing graph data present in such an incremental reconstruction scenario.
   Although our proposed method is designed to run on submaps of the scene, we show it also transfers to entire 3D scenes. Experiments show that our approach outperforms 3D scene graph prediction methods by a large margin and its accuracy is on par with other 3D semantic and panoptic segmentation methods while running at \methodHz Hz.
\end{abstract}%

\section{Introduction}
High-level scene understanding is a fundamental task in computer vision required for many applications in fields such as robotics and augmented or mixed reality. Boosted by the availability of inexpensive depth sensors, real-time dense SLAM algorithms~\cite{Newcombe2011_KinectFusion, Keller2013, Niessner2013, Whelan2015} and large scale 3D datasets~\cite{Dai2017_scannet, Wald2019RIO}, the research focus has shifted from reconstructing the 3D scene geometry to enhancing the 3D maps with semantic information about scene components. Several methods have deployed a neural network to process a complete 3D scan of a scene~\cite{Dai2017_scannet, Engelmann_2017_ICCV, Rethage2018, Ruizhongtai2017, Lahoud2019_MTML, Huang2019_TextureNet, Hou2019_SIS, han2020occuseg, Engelmann2020_3DMPA}. However, these all require 3D geometry as prior information and they typically operate in an offline fashion, \ie without satisfying real-time requirements, which are fundamental for many real-world applications. Real-time scene understanding that incrementally built 3D scans poses important challenges such as handling partial, incomplete, and ambiguous scene geometry where object shapes may change dramatically over time. Learning a robust 3D feature that can cope with this variability is difficult. Furthermore, fusing multiple, potentially contradictory network predictions to ensure consistency in the global map, is also challenging. Recently, in the image domain, semantic scene graphs have been used to derive relationships among scene entities~\cite{li2017scene, Xu2017, Newell2017, Yang2018, gu2019scene}. Scene graphs demonstrated to be a powerful abstract representation for scene understanding. Being compact and explicit, they are beneficial for complex tasks such as image captioning~\cite{Xu2015, Karpathy2015}, generation~\cite{Johnson2018}, manipulation~\cite{dhamo2020semantic} or visual questioning and answering~\cite{Teney2017}. For this reason, recent works have explored scene graph prediction from entire 3D scans in an offline manner~\cite{Wald2020_3dssg, Armeni2019_3dsg}. Furthermore, building up semantic graph maps \textit{online} is a major challenge, requiring not only to efficiently detect semantic instances in the scene but also to robustly estimate predicates between them, while dealing with partial and incomplete 3D geometry.\par%
In this work, we propose a real-time method to incrementally build, in parallel to 3D mapping, a globally consistent semantic scene graph, as shown in Fig.~\ref{fig:teaser}. Our approach relies on a geometric segmentation method~\cite{Tateno2015} and a novel inductive graph network, which handles missing edges and nodes in partial 3D point clouds. Our scene nodes are geometric segments of primitive shapes. Their 3D features are propagated in a graph network that aggregates features of neighborhood segments. Our method predicts scene semantics and identifies object instances by learning relationships among clusters of over-segmented regions. Towards this end, we propose to learn additional relationships, referred to as \texttt{same part} in an end-to-end manner.\par%
The main contributions of this work can be summarized as follows: %
(1) We propose the first online 3D scene graph prediction, \ie incrementally fusing predictions from currently observed sub-maps into a globally consistent semantic graph model. %
(2) Due to a new relationship type, nodes are merged into 3D instances, resembling panoptic segmentation.%
(3) We introduce a novel attention method that can handle partial and incomplete 3D data, as well as highly dynamic edges, which is required for incremental scene graph prediction. %
Our experiments show that we outperform 3D scene graph prediction and achieve on par performance on 3D semantic and instance segmentation benchmarks while running in \methodHz Hz.\par%
\begin{figure*}[th]
    \centering
    \includegraphics[width=\textwidth]{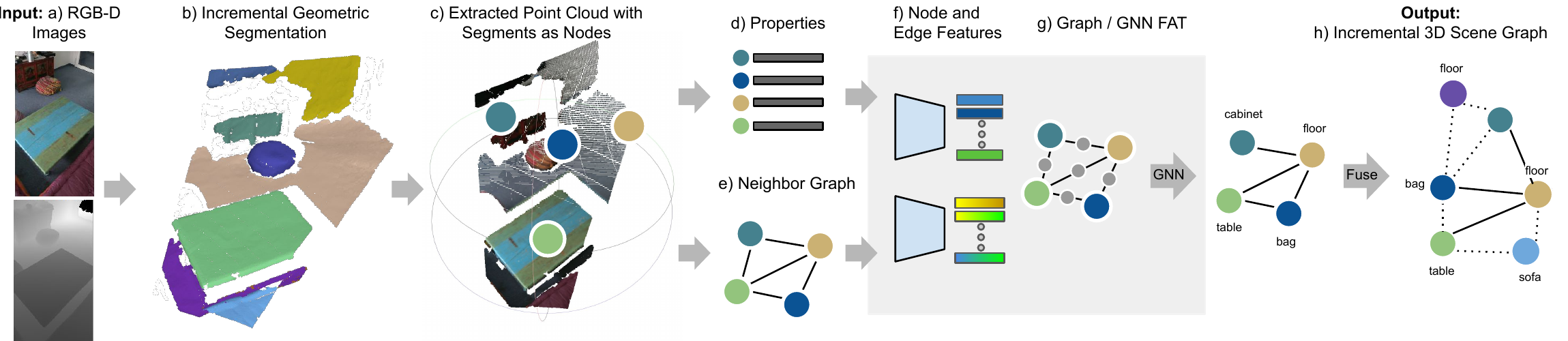}
    \caption{Overview of the proposed SceneGraphFusion framework. Our method takes a stream of RGB-D images a) as input to create an incremental geometric segmentation b). Then, the properties of each segment and a neighbor graph between segments are constructed.
    The properties d) and neighbor graph e) of the segments that have been updated in the current frame c) are used as the inputs to compute node and edge features f) and to predict a 3D scene graph g). Finally, the predictions are h) fused back
into a globally consistent 3D graph.    
    }
    \label{fig:system_overview}
\end{figure*}%
\begin{figure}
    \centering
    \includegraphics[width=\columnwidth]{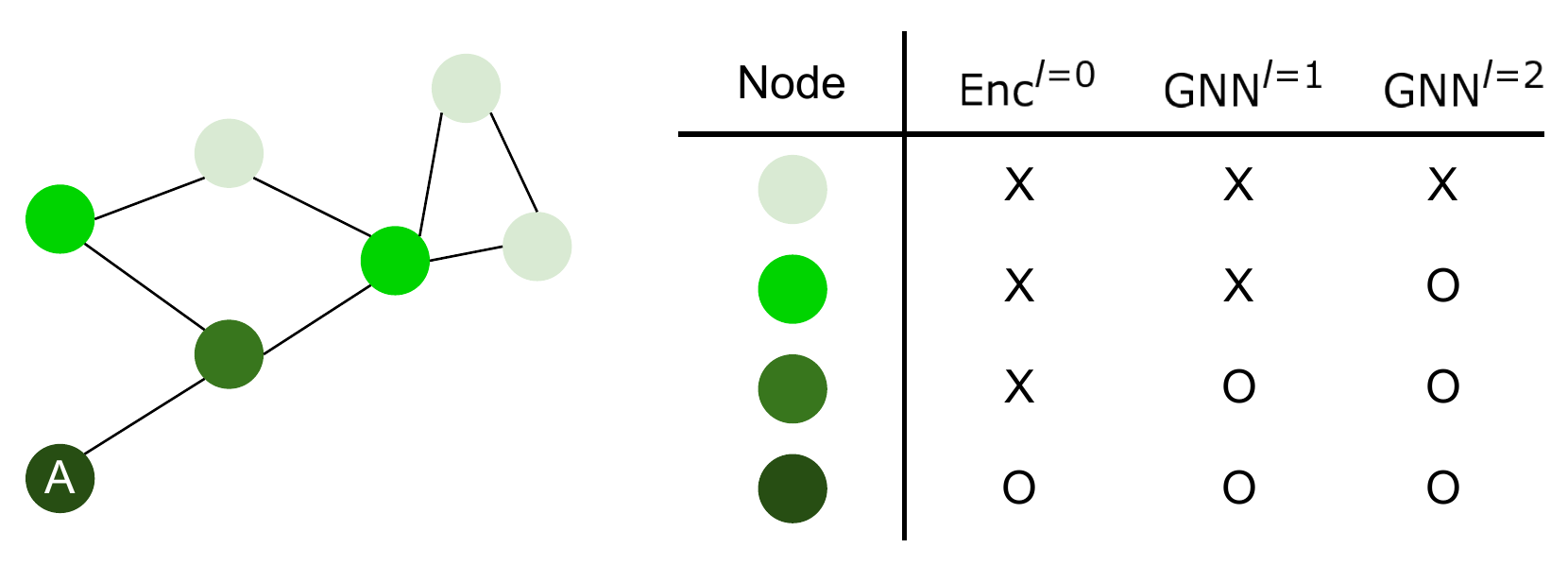}
    \caption{A representation of our efficient graph update strategy. Given a network with a basic encoder (\({\rm Enc}^{l=0}\)) and two message passing layers (\({\rm GNN}^{l=1}, {\rm GNN}^{l=2}\)), by storing different layers of features separately. When node A is updated, we can reuse the lower-layer features from other nodes without recomputing them. We visualize the operations needed at each node with color. 
    }
    \label{fig:updategraph}
\end{figure}%
\section{Related Work}%
\subsection{Semantic SLAM}%
Several 3D scene understanding methods leverage deep learning to perform either semantic segmentation~\cite{Dai2017_scannet, Engelmann_2017_ICCV, Rethage2018, Ruizhongtai2017, Huang2019_TextureNet}, or instance segmentation/object detection~\cite{Hou2019_SIS, Qi2019_VoteNet, Lahoud2019_MTML, Engelmann2020_3DMPA} from the complete 3D volume or point cloud of the scene. Conversely, incremental  semantic SLAM approaches 
do not assume a full 3D scan to be available, instead directly operate on the incoming frames of RGB(-D) sequences~\cite{McCormac2017_semanticfusion, Valentin2015, Rosinol2020}. Such methods simultaneously carry out a 3D reconstruction of the scene, while extracting the corresponding semantics of the currently observed surface. To this end, some incremental methods transfer image predictions from a convolutional neural network (CNN) to 3D, passing the data from the image to the 3D reconstruction~\cite{McCormac2017_semanticfusion}. \cite{Tateno2017} propose a monocular approach that constructs the 3D geometry from a depth prediction, rather than a depth image. These incremental approaches often require a sophisticated fusion and/or a regularization method to deal with multiple, potentially contradictory, predictions, and to handle spatial and temporal consistency~\cite{McCormac2017_semanticfusion, Gaku2019_panopticfusion, Tateno2017}. Other approaches fuse the 2D image and 3D reconstruction~\cite{Jiazhao2020_fusionaware}. These semantic SLAM methods such as SemanticFusion~\cite{McCormac2017_semanticfusion}, ProgressiveFusion~\cite{pham2019real} or FusionAware~\cite{Jiazhao2020_fusionaware} are able to reconstruct 3D semantic scene maps in real-time, but are not able to differentiate between individual object instances. Object-level SLAM approaches focus on object instances while sometimes requiring prior knowledge of the scene such as an object database or semantic class annotations~\cite{SalasMoreno2013, Tateno2016, McCormac2018_fusionplusplus, Furrer2018, Grinvald2019}.\par%
Segmentation techniques are often used to reduce data complexity and meet the required runtime on limited resources. Several methods~\cite{Tateno2016, Li2016, Nakajima2018, Wald2018, Grinvald2019} incorporate the efficient incremental segmentation method proposed in~\cite{Tateno2015} to perform online scene understanding.\par%
Semantic SLAM methods achieve great performance in computing a semantic or object-level representation but neither focus on semantic scene graphs nor on semantic relationships between object instances.\par%
\subsection{Scene Graphs for Images and 3D Data}%
Graph Neural Networks (GNNs) have recently emerged as a popular inference tool for many challenging tasks~\cite{Ashish2017_attention, Qi2018_attentive, Manessi2018, Velickovi2018_GAT, Cosmo2020, Sarlin2020_superglue}. In particular, GNNs have been proposed to infer scene graphs from images~\cite{Yang2018, Qi2018_attentive}, where scene entities are the nodes of the graph, \eg object instances. Scene graph prediction goes beyond instance segmentation by adding relationships between instances. Although scene graphs are adapted from computer graphics, their semantic extension has become an important research area in computer vision. Since the introduction of a large scale 2D scene graph dataset~\cite{Krishna2017_visualgenome}, several graph prediction methods have been proposed focusing on message passing with recurrent neural networks~\cite{Xu2017}, iterative statistical optimization~\cite{Dai2017_DeepRel} or methods to handle limited data~\cite{Chen2019, Dornadula2019}. Furthermore, recent datasets with 3D semantic scene graph annotations have been proposed~\cite{Gay2018_visual, Armeni2019_3dsg, Wald2020_3dssg}, alongside with 3D graph estimation methods.~\cite{Wald2020_3dssg} predict semantic scene graphs from a ground truth class-agnostic segmentation of the 3D scene.~\cite{Gay2018_visual} use an object detector on a sequence of images to construct 3D quadrics -- their object representation of choice. The geometric and visual features are then processed with a recurrent neural network.~\cite{Armeni2019_3dsg} use mask predictions and a multi-view regularization technique on sampled images to compute relationships derived from detected object instances. They construct a 3D scene graph of a building that includes object semantics, rooms and cameras, as well as the relationships between these entities.~\cite{Rosinol2020_dynamic} extended this model to include dynamic scene entities, e.g. humans. Nevertheless, all of these methods work offline and expect the reconstructed 3D scene as an input.\par%
\noindent
Similarly to~\cite{Wald2020_3dssg}, we predict graphs of semantic nature, but in contrast to~\cite{Wald2020_3dssg}, our graph prediction does not require any prior scene knowledge and is able to segment instances and their semantic information, as well as their relationships, in real-time, while the scene is being reconstructed.%
%
\section{Incremental 3D Scene Graph Framework}%
Fig.~\ref{fig:system_overview} illustrates the pipeline of our SceneGraphFusion framework. Our system consists of two separate cores: a reconstruction and segmentation pipeline adapted from~\cite{Tateno2015} (Sec.~\ref{sec:inseg}), and a scene graph prediction network (SPN) (Sec.~\ref{sec:gnn}). Our system takes a sequence of RGB-D frames with associated poses as input to reconstruct a segmented map of the scene, while estimating a neighbor graph and properties of each segment. Then, a subset of the neighbor graph and the properties of the segments that have been recently observed are fed into our graph network to predict node and edge semantics. Finally, the predictions are fused into the globally consistent 3D scene graph. To maintain real-time performance of our system, we separate the scene graph prediction process into a different thread. The 3D scene graph is asynchronously predicted and fused from the reconstruction pipeline. Our semantic scene graph \(\mathcal{G}\) consists of a set of tuples \((\mathcal{V}, \mathcal{E})\) with nodes \(\mathcal{V}\) and edges \(\mathcal{E}\). Nodes represent segments with their object categories, and edges represent the semantic relationships (predicates) between nodes, such as {\tt standing on} and {\tt attached to}.\par%
\subsection{Scene Reconstruction with Property Building}\label{sec:inseg}%
An incremental and computationally efficient method of estimating instances is required to enable construction of a scene graph in real-time. We use the incremental geometrical segmentation method in~\cite{Tateno2015} to build a globally consistent segmentation map, and incorporate it with our online property update and neighbor graph building.\par%
\paragraph{Geometric Segmentation and Reconstruction.} Given input RGB-D frames and associated poses, the incremental segmentation algorithm generates a global 3D segmentation map, shown in Fig.~\ref{fig:system_overview}b, by performing incremental segmentation on top of a dense reconstruction algorithm. The 3D segmentation map consists of a set of segments \(\mathbf{S} = \{\mathbf{s}_{1}, \dots, \mathbf{s}_{n}\}\). Each segment stores a set of 3D points $\mathbf{P}_i$ where each point has a 3D coordinate a normal and a color. Our map is updated at every new frame, by adding new segments and merging or removing old ones.\par%
\paragraph{Segment Properties.} In addition to segment reconstruction, we compute segment properties (see Fig.~\ref{fig:system_overview}d) to describe a segment shape, \ie centroid \( \overline{\mathbf{p}}_i  \in \mathbb{R}^3\), standard deviation of the position of points \(\bm{\sigma}_i \in \mathbb{R}^3\), size of the axis-aligned bounding box \( \mathbf{b}_i =  \left(b_x, b_y, b_z\right) \in \mathbb{R}^3\), maximum length \( l_i = \max \left ( b_x, b_y, b_z  \right ) \in \mathbb{R}\) and bounding box volume \( \nu_i = b_x \cdot b_y  \cdot b_z  \in \mathbb{R}\). Reconstructing the segments in this incremental manner allows us to update the properties of each node efficiently. These properties are updated by checking every modification of the points in the segment.\par%
\paragraph{Neighbor Graph.}%
Additionally, we construct a neighbor graph having nodes as segments and edges as the connection between the segments, as depicted in Fig.~\ref{fig:system_overview}e. To find the adjacent segments, we compute the distances between all the combinations of the bounding box of the segments. The segment pairs where the distance is less than a certain threshold are added as edges (we use 0.5 meters as a proximity threshold in our experiments).\par%
\subsection{Prediction with Graph Structure}\label{sec:prediction}%
Next, we feed the segments properties and the neighbor graph to our graph network to predict the segment label and predicate on each segment and edge, shown in Fig.~\ref{fig:system_overview}f-g. The detailed description of our graph network architecture can be found in Sec.~\ref{sec:gnn}. Since our segment reconstruction process is incremental, only those segments that are currently observed in the input frame are updated. Therefore, we only feed a subset of the segments and the neighbor graph which consists of the segments that have been updated in recent frames, this improving scalability and efficiency. To identify the newly updated segments, we store the segment size and timestamp whenever the segment is fed into the network. If the segment size changes more than 10\%, or the segment has not been updated for 60 frames, they are flagged and fed into the network. Segments are continuously observed and outdated segments and their neighbors are extracted and processed with our graph neural network. For the sake of efficiency we store all features computed from our SPN in our neighbor graph. According to the message passing process in GNN~\cite{Gilmer2017neural}, when a lower-layer feature of a node is updated, only the higher-layer features of this node, its direct neighbors, and their edge features are affected. This allows us to re-use previously computed features, as shown in Fig.~\ref{fig:updategraph}, and greatly improve prediction efficiency and scalability.\par%
\subsection{Temporal Scene Graph Fusion}\label{sec:fusion}%
Finally, the predicted semantics of nodes and edges in the neighbor graph are fused into a globally consistent semantic scene graph, depicted in Fig.~\ref{fig:system_overview}h. Due to the incremental nature of our method, as described in Sec.~\ref{sec:prediction}, the semantics of each segment and edge are predicted multiple times, resulting in potentially contradictory outcomes. To handle this, we apply a running average approach~\cite{curless1996volumetric} to fuse the predictions of the same segment or edge. For each segment and edge in our neighbor graph, we store a weight \(w\) and a probability \(\mu\) for each class or predicate prediction. Given a new prediction with probability \(\mu^{t}\) at time \(t\), we update the previously stored weight \(w^{t-1}\) and probability \(\mu^{t-1}\) as %
\begin{gather}
    \mu^{t} = \frac{\mu^{t}\cdot w^{t} + \mu^{t-1}\cdot w^{t-1}}{w^{t}+w^{t-1}},\\
    w^{t} = \min\left( w_{max}, w^{t} + w^{t-1} \right ),
\end{gather}
where \(w_{max} = 100\) is the maximum weight value.
Importantly, since our framework predicts semantics at segment level, we are able to store and preserve the whole label probability distribution using a much smaller memory footprint compared to point-level methods~\cite{McCormac2017_semanticfusion}.\par%
%
\section{Scene Graph Prediction}\label{sec:gnn}%
The use of segments obtained by the geometric segmentation method requires the design of a robust feature, since the shape of each segment is usually incomplete and relatively simple, and changes overtime during reconstruction. The feature of each segment can be enhanced with neighbor information by using a GNN. However, the number of neighbors of each segment changes over time, posing a serious challenge for the training process.\par%
Dealing with dynamic nodes and edges in a GNN is known as inductive learning. Existing methods focus mainly on how to spread attention across all the neighbors~\cite{Ashish2017_attention, Velickovi2018_GAT}, or estimate the attention between nodes~\cite{Cosmo2020}. However, in either case, a missing edge still affects all the aggregated messages. To deal with this problem, we propose a novel feature-wise-attention (FAT), that re-weights individual latent features at each target node. By applying a max function on this re-weighted embedding, this strategy yields aggregated features that are less affected by missing neighboring points.\par%
\subsection{Network Architecture}\label{sec:network_architecture}%
The network architecture of our framework is shown in Fig.~\ref{fig:system_overview}f-g (grey box). Our architecture is inspired from~\cite{Wald2020_3dssg}, with modification of some major components. %
Given a) a set of segments, b) the properties of each segment, and c) a neighbor graph, our network outputs a semantic scene graph by predicting class and predicate for each segment and edge respectively.\par%
\paragraph{Node Feature.} The point cloud \(\mathbf{P}_i\) of each segment \(\mathbf{s}_i\) is encoded with a PointNet~\cite{Ruizhongtai2016_pointnet} \(f_p\left(\cdot\right)\) into a latent feature that represents the primitive shape of each segment. We concatenate the spatial invariant properties described in Sec.~\ref{sec:inseg}, \ie standard deviation \(\bm{\sigma}_i\), log of bounding box size \(\mathbf{b}_i\), length \(l_i\), and volume \(\nu_i\), with \(f_p(\mathbf{P}_i)\) to handle the scale insensitive limitation caused by normalization of the input points on the unit sphere such that%
\begin{equation}
    \mathbf{v}_{i} = [f_p\left(\mathbf{P}_i\right), \bm{\sigma}_i, \ln{ \left  ( \mathbf{b}_{i} \right ) },  \ln{\left( \nu_i \right)} \ln{ \left  ( l_i \right ) }],
\end{equation}%
where \([\cdot]\) denotes a concatenation function.\par%
\paragraph{Edge Feature.} The visual features of the edges are computed with the properties of the connected segments. Given an edge between a source node \(i\) and a target node \(j\) where \(j \neq i\), the edge visual feature \(\mathbf{e}_{ij}\) is computed such that%
\begin{gather}
    \textstyle
    \mathbf{{r}}_{ij} = [\overline{\mathbf{p}}_i-\overline{\mathbf{p}}_j, \bm{\sigma}_i-\bm{\sigma}_j, \mathbf{b}_i-\mathbf{b}_j  , \ln{ \left( \frac{l_{i}}{l_{j}} \right) } ,\ln{\left( \frac{\nu_i}{\nu_j} \right) }  ], \\
    \mathbf{e}_{ij} = g_s \left( \mathbf{r}_{ij} \right),
\end{gather}%
where \(g_s \left(\cdot\right) \) is a multi-layer perception (MLP) projecting the paired segment properties into a latent space.\par%
\paragraph{GNN Feature.} After the initial feature embedding on nodes and edges, we propagate the features using a GNN with 2 message passing layers to enhance the features by enclosing the neighborhood information. Our GNN updates both node and edge features in each message passing layer \(\ell\). In each layer, the node and \(\mathbf{v}^{\ell}_{i}\) and edge features \(\mathbf{e}^{\ell}_{ij}\) are updated as follows: %
\begin{gather}
    \mathbf{v}^{\ell+1}_{i} = g_{v}\left( [\mathbf{v}^{\ell}_{i},
    \max_{j\in \mathcal{N}\left(i\right)}\left( \text{FAN}\left( \mathbf{v}^{\ell}_{i},\mathbf{e}^{\ell}_{ij},\mathbf{v}^{\ell}_{j} \right) \right) ]\right),\\
    \mathbf{e}^{\ell+1}_{ij} = g_{e} \left([ \mathbf{v}^{\ell}_{i}, \mathbf{e}^{\ell}_{ij}, \mathbf{v}^{\ell}_{j} ]\right),
\end{gather}%
where \(g_{v}\left(\cdot\right)\) and \(g_{e}\left(\cdot\right)\) are MLPs, \(\mathcal{N}\left(i\right)\) is the set of neighbors indices of node \(i\), and FAN\(\left(\cdot\right)\) is the proposed feature-wise attention network, which is detailed in Sec.~\ref{sec:attention}.\par%
\paragraph{Class Prediction and Losses.} Finally, the node class and the edge predicate are predicted by means of two MLP classifiers. Similarly to~\cite{Wald2020_3dssg}, our network can be trained end-to-end with a joint cross entropy loss, for both, object \(\mathcal{L}_{obj}\) and predicates \(\mathcal{L}_{pred}\).\par%
%
\subsection{Feature-wise Attention}\label{sec:attention}
Our feature-wise attention (FAT) module takes as input a query \(\mathbf{Q}\) of dimensions \(d_q\) and targets \(\mathbf{T}\) of dimensions \(d_{\tau}\). %
It estimates a weight distribution of dimensions \(d_{\tau}\) by using a MLP \(g_{a}\left(\cdot\right)\) with a softmax operation to normalize and distribute the weight. Then, the attention is calculated by element-wise multiplication of the weight matrix and the target \(\mathbf{T}\),
\begin{equation} \label{eq:attention}
    \text{FAT}(\mathbf{Q}, \mathbf{T}) = \text{softmax} \left(g_{a}(\mathbf{Q})\right) \odot \mathbf{T},
\end{equation}%
where \(\odot\) denotes element-wise multiplication.\par%
The use of softmax across the entire target dimension \(d_\tau\) gives us a single weight matrix across all feature dimensions. We employ a multi-head approach as in~\cite{Ashish2017_attention, Sarlin2020_superglue} to allow a more flexible attention distribution. %
The input feature dimension of \(\mathbf{Q}\) and \(\mathbf{T}\) are divided into \(h\) heads \(\mathbf{Q}=\left[\mathbf{q}_1, \dots,\mathbf{q}_h\right]\) and \(\mathbf{T}=\left[\bm{\tau}_1, \dots,\bm{\tau}_h\right]\) with \(\mathbf{q}_i\in \mathbb{R}^{d_q/h}\) and \(\bm{\tau}_i\in \mathbb{R}^{d_{\tau}/h}\). For each head, the same attention function as in equation ~(\ref{eq:attention}) is applied, then the values from each head are concatenated back to the dimensions \(d_\tau\) to obtain the multi-head attention:%
\begin{equation}\label{eq:attention_full}%
\text{MFAT}\left(\mathbf{Q},\mathbf{T}\right) = \big[\text{FAT}(\mathbf{q}_{i},\bm{\tau}_{i})\big]_{i=1}^{h}.
\end{equation}%
Unlike scaled dot-product attention~\cite{Ashish2017_attention}, our approach does not distribute across edges. Instead, it learns to spread the attention across the feature dimensions of each target node. %
Towards this end, we design a feature-wise attention network (FAN). %
Given a source node feature \(\mathbf{v}_{i}\), an edge feature \(\mathbf{e}_{ij}\), and a target node feature \(\mathbf{v}_{j}\), we compute the weighted message as%
\begin{equation}\label{eq:fan}%
\begin{aligned}
    \text{FAN}&\left( \mathbf{v}_{i}, \mathbf{e}_{ij}, \mathbf{v}_{j} \right)=\\
    & \text{MFAT} \left( \left[ \hat{g}_{q}\left(\mathbf{v}_{i}\right), \hat{g}_{e}\left(\mathbf{e}_{ij}\right) \right], \hat{g}_{\tau}\left(\mathbf{v}_{j}\right) \right),
\end{aligned}%
\end{equation}%
where \(\hat{g}_{q}\left( \cdot \right), \hat{g}_{e}\left( \cdot \right), \hat{g}_{\tau}\left( \cdot \right)\) are single layer perceptrons to map \(\mathbf{v}_{i}, \mathbf{e}_{ij}, \mathbf{v}_{j}\) to dimensions \(\frac{d_q}{2}, \frac{d_q}{2}, d_{\tau}\) respectively.\par%
%
\section{Data Generation}\label{sec:data_gen}%
\begin{figure}[t]
    \centering
    \includegraphics[width=\columnwidth]{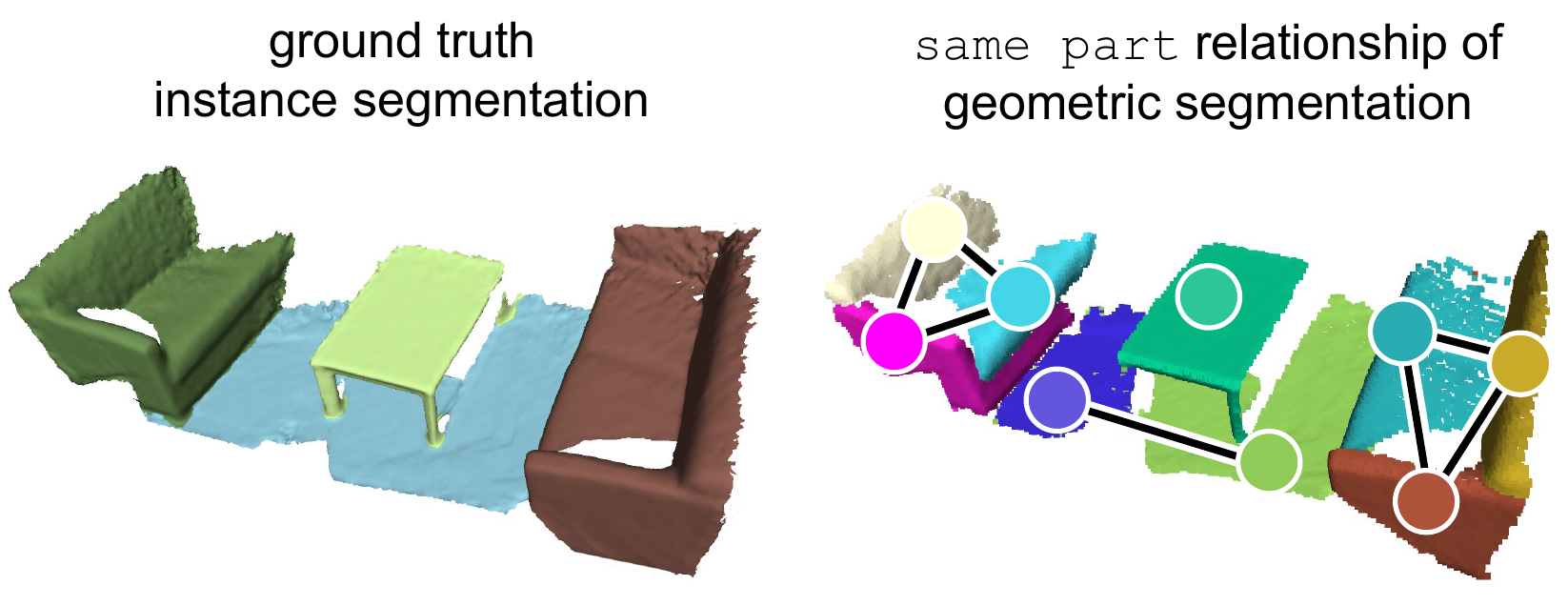}
    \caption{The \texttt{same part} relationship is generated between the segments corresponding to the same object instance.}
    \label{fig:samepart}
\end{figure}

We introduce a \texttt{same part} relationship to allow our network to cluster segments from the same object, enabling instance-level object segmentation on the over-segmented map. %
We generate training and testing data using the estimated segmentation and the ground truth instance annotations.
Given a scene segmented by a geometrical segmentation method and its corresponding ground truth provided as object instance annotations, we find the best match between each estimated segment and the ground truth objects via nearest neighbor search. %
In particular, the best match is obtained by maximizing the area of intersection between the given segment and the ground truth objects. We reject matches where the area of intersection is less than $50\%$ of the segment surface. %
In addition, we consider a valid match only if its corresponding segment does not cover any other ground truth objects by more than $10\%$ of their area. In the case of multiple segments corresponding to the same object instance, we add the \texttt{same part} relationship between all of them, as shown in Fig.~\ref{fig:samepart}. Finally, if ground truth relationships exist on that object instance, they are inherited by all the segments.\par%
%
\section{Evaluation}\label{sec:evaluation}%
In Sec.~\ref{sec:eval_scene_graph_pred} we evaluate our scene graph prediction on 3DSSG~\cite{Wald2020_3dssg}. The performance of our method is reported on full scenes given ground truth instances and second with geometric segments. We then show how relationships/graphs help with object prediction. In Sec.~\ref{sec:semantic_panoptic_segmentation} we focus on the by-product of our method, panoptic segmentation by reporting segmentation scores on ScanNet~\cite{Dai2017_scannet}. Finally, in Sec.~\ref{sec:runtime} we provide a runtime analysis of our method compared to other incremental semantic segmentation approaches.\par%
\subsection{Semantic Scene Graph Prediction}\label{sec:eval_scene_graph_pred}%
Following the evaluation scheme in~\cite{Wald2020_3dssg}, we separately report relationship, object, and predicate prediction accuracy with a top-n evaluation metric. Following~\cite{Yang2018}, the relationship score is the multiplication of the object, subject, and predicate probability. Object and predicate metrics are calculated directly with the respective classification scores.\par%
\paragraph{Ground Truth Instances.} In Tbl.~\ref{tab:3dssg_full_160_26} we report the 3D scene graph prediction accuracy independently from the segmentation quality. The evaluation was conducted on the full 3D scene with the class-agnostic ground truth segmentation, as carried out in~\cite{Wald2020_3dssg}. We followed the data split proposed in 3DSSG with 160 object classes and 26 different relationships. Our method outperforms~\cite{Wald2020_3dssg} with a significant margin of +0.45 / +0.21 (R@50 / R@100) for relationship prediction due to small improvements in predicate and object classification. Note that our method can run offline on the pre-computed 3D data -- as done here -- but is also able to handle partial and incomplete shapes in an incremental online setup which is analyzed in following paragraph.\par%
\setlength{\tabcolsep}{2pt}
\begin{table}[t]
    \begin{center}
    \resizebox{\columnwidth}{!}{
    \begin{tabular}{p{2.2cm}|p{0.9cm}P{0.9cm} p{0.1cm} P{0.9cm}P{0.9cm} p{0.1cm} P{0.9cm}P{0.9cm}} & \multicolumn{2}{l}{Relationship} & & \multicolumn{2}{c}{Object} & & \multicolumn{2}{c}{Predicate} \\
&  R@50 & R@100 & & R@5 & R@10 & & R@3 & R@5  \\
\hline
Baseline~\cite{Wald2020_3dssg} & 0.39 &  0.45 & & 0.66 & 0.77 & & 0.62 & 0.88 \\
3DSSG~\cite{Wald2020_3dssg} & 0.40 & 0.66 & & 0.68 & 0.78 & & 0.89 & 0.93 \\
Ours & \textbf{0.85} & \textbf{0.87} & & \textbf{0.70} & \textbf{0.80} & & \textbf{0.97} & \textbf{0.99} \\ 
    \end{tabular}
    }
    \end{center}
    \caption{Evaluation of the scene graph prediction task on 3RScan/3DSSG~\cite{Wald2020_3dssg} with 160 objects and 26 predicate classes. The experiments were conducted on the complete 3D data.
    }
    \label{tab:3dssg_full_160_26}
\end{table}%

\paragraph{Geometric Segments.}%
\setlength{\tabcolsep}{2pt}
\begin{table}[t]
    \begin{center}
    \resizebox{\columnwidth}{!}{
    \begin{tabular}{llc|cccccc} & & & \multicolumn{2}{c}{Relationship} & \multicolumn{2}{c}{Object} & \multicolumn{2}{c}{Predicate} \\
          & Method (Attention) & & R@1 & R@3 &  R@1 & R@3 & R@1 & R@2 \\
         \hline
        \textcircled{\raisebox{-0.4pt}{\scriptsize 1\normalsize}} & 3DSSG (none) & 
        \((f)\) & 0.38 & 0.59 & 0.61 & 0.85 & 0.83 & 0.98 \\ 
         \textcircled{\raisebox{-0.4pt}{\scriptsize 2\normalsize}} & Ours (none) & 
         \((f)\)  & 0.41 & 0.62 & 0.62 & 0.88 & 0.84 & 0.98 \\ 
         \textcircled{\raisebox{-0.3pt}{\scriptsize 3\normalsize}} & Ours (GAT) & 
         \((f)\) & 0.12 & 0.22 & 0.25 & 0.64 & 0.85 & 0.98 \\ 
         \textcircled{\raisebox{-0.3pt}{\scriptsize 4\normalsize}} & Ours (SDPA) & \((f)\) & 0.39 & 0.62 & 0.62 & 0.87 & 0.85 & 0.98 \\ 
         \textcircled{\raisebox{-0.4pt}{\scriptsize 5\normalsize}} & Ours (FAT) & \((f)\) & \textbf{0.55} & \textbf{0.78} & 0.75 & 0.93 & \textbf{0.86} & 0.98 \\ 
         \textcircled{\raisebox{-0.4pt}{\scriptsize 6\normalsize}} & Ours (FAT) &  \((i)\) & 0.51 & 0.67 & 0.78 & \textbf{0.94} & 0.77 & 0.98 \\
         \textcircled{\raisebox{-0.4pt}{\scriptsize 7\normalsize}} & Ours Fusion (FAT) & \((i)\) & 0.52 & 0.70 & \textbf{0.79} & \textbf{0.94} & 0.78 & 0.98 \\
    \end{tabular}
    }
    \end{center}
    \caption{Evaluation of the semantic scene graph prediction on geometric segments of 3RScan/3DSSG~\cite{Wald2020_3dssg} with 20 objects and 8 predicate classes. \((f)\) indicates a prediction on the full 3D scene while \((i)\) is the incremental result from the RGB-D sequence.}
\label{tab:3dssg_inseg_20_8}
\end{table}%
In Tbl.~\ref{tab:3dssg_inseg_20_8} we compare the performance of incremental \textcircled{\raisebox{-0.4pt}{\scriptsize 6\normalsize}}-\textcircled{\raisebox{-0.4pt}{\scriptsize 7\normalsize}} and full scene graph prediction \textcircled{\raisebox{-0.4pt}{\scriptsize 5\normalsize}} based on our geometric segmentation. \textcircled{\raisebox{-0.4pt}{\scriptsize 6\normalsize}} is slightly worse than \textcircled{\raisebox{-0.4pt}{\scriptsize 5\normalsize}} but generates predictions on the fly. %
Our proposed fusion \textcircled{\raisebox{-0.4pt}{\scriptsize 7\normalsize}} improves the performance further. 
Tbl.~\ref{tab:3dssg_inseg_20_8} additionally shows that we outperform 3DSSG~\cite{Wald2020_3dssg} with a small margin without any attention method and with a large margin when using our proposed feature-wise attention, FAT \textcircled{\raisebox{-0.4pt}{\scriptsize 5\normalsize}}. %
FAT \textcircled{\raisebox{-0.4pt}{\scriptsize 5\normalsize}} also outperforms other attention mechanisms GAT~\cite{Velickovi2018_GAT} \textcircled{\raisebox{-0.4pt}{\scriptsize 3\normalsize}} and SDPA~\cite{Ashish2017_attention} \textcircled{\raisebox{-0.4pt}{\scriptsize 4\normalsize}} for 3D semantic scene graph prediction. The input of the methods is either the full 3D scene \textcircled{\raisebox{-0.4pt}{\scriptsize 1\normalsize}}-\textcircled{\raisebox{-0.4pt}{\scriptsize 6\normalsize}}, processed offline \((f)\) or a stream of RGB-D images processed incrementally \((i)\), \textcircled{\raisebox{-0.4pt}{\scriptsize 6\normalsize}}, \textcircled{\raisebox{-0.4pt}{\scriptsize 7\normalsize}}. 
For these experiments, we first acquired the geometric segmentation~\cite{Tateno2017} from the RGB-D sequences of 3RScan~\cite{Wald2019RIO}. The final training data was generated with the pipeline described in Sec.~\ref{sec:data_gen}. 
We trained the networks with 20 NYUv2~\cite{Silberman2012} object classes used on the ScanNet~\cite{Dai2017_scannet} benchmark. Furthermore, only support predicates are used and relationships with too few occurrences are ignored. 
This leads to 8 predicates, including the \texttt{same part} relationship which we added in the data generation process. 
More details on the training setup and chosen hyper-parameters used in this experiment can be found in the supplementary material. %
A qualitative result of our graph prediction is shown in Fig.~\ref{fig:evalitative_evalation}, more examples can also be found in the supplementary.\par%
\paragraph{Predicate Influence on Object Classification.} To verify if learning inter-instance relationship improves object classification, we train our network without the predicate loss. Tbl.~\ref{tab:eva_without_predicate_loss} shows that object classification indeed benefits from joint relationship prediction.\par%
\setlength{\tabcolsep}{2pt}
\begin{table}[t]
    \begin{center}
    \resizebox{\columnwidth}{!}{
    \begin{tabular}{p{2.9cm}|p{0.8cm}P{0.8cm} p{0.1cm} P{0.8cm}P{0.8cm} p{0.1cm} P{0.8cm}P{0.8cm}} &  \multicolumn{2}{l}{Relationship} &  & \multicolumn{2}{c}{Object} & & \multicolumn{2}{c}{Predicate} \\
          & R@1 & R@3 & & R@1 & R@3 & & R@1 & R@2  \\
         \hline
         Ours without $\mathcal{L}_{pred}$ & 0.26 & 0.36 & & 0.62 & 0.87 & & 0.59 & 0.75\\
         Ours with $\mathcal{L}_{pred}$& \textbf{0.55} & \textbf{0.78} & & \textbf{0.75} & \textbf{0.93} & & \textbf{0.86} & \textbf{0.98}\\
    \end{tabular}
    }
    \end{center}
    \caption{Ablation Study: Comparison of training with and without predicate loss $\mathcal{L}_{pred}$ on 3RScan/3DSSG~\cite{Wald2020_3dssg} with 20 object and 8 predicate classes. Note that the comparison is based on graphs computed from the full 3D scene \((f)\).}
\label{tab:eva_without_predicate_loss}
\end{table}%
\begin{figure*}[t]
    \centering
    \includegraphics[width=\textwidth]{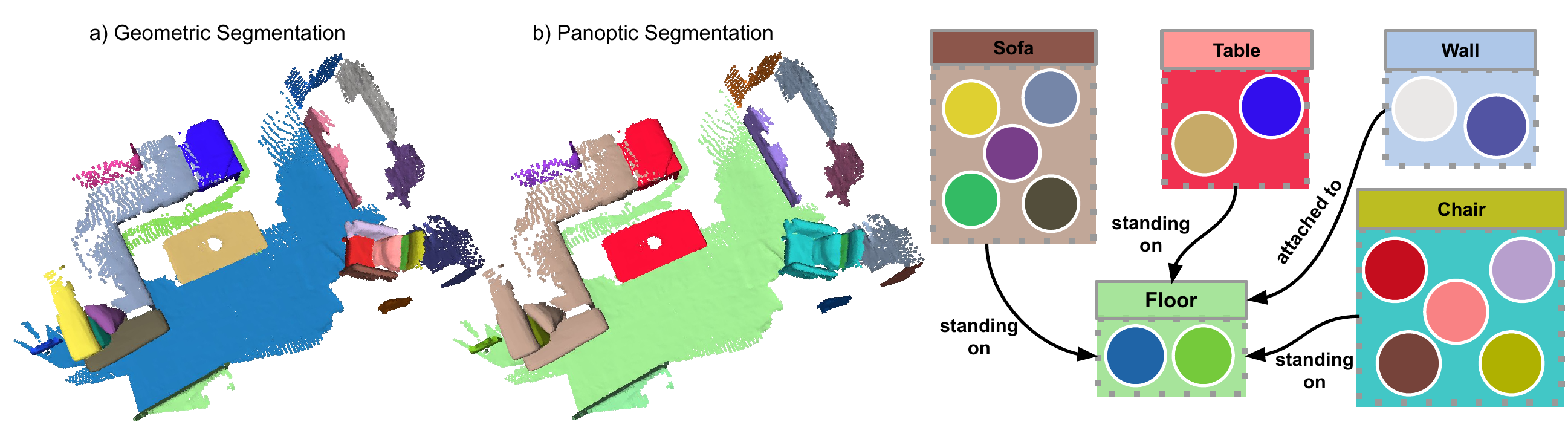}
    \caption{Qualitative evaluation of our incremental graph prediction. Node (circle) colors represent geometric segments shown on the left, the corresponding predicted semantics (panoptic segmentation) is visualized in the center, corresponding to the boxes in the right. For visualization purposes, we only show the biggest segments and filter out small ones.}
    \label{fig:evalitative_evalation}
\end{figure*}%
%
\setlength\dashlinedash{1.2pt}
\setlength\dashlinegap{1.5pt}
\setlength{\tabcolsep}{0.8pt}
\begin{table}[t]
    \begin{center}
    \resizebox{\columnwidth}{!}{
    \begin{tabular}{p{3.5cm}P{1.0cm}P{1.2cm}P{1.2cm}P{1.2cm}}
     & Metric & All & Things & Stuff\\ 
    \hline
    PanopticFusion~\cite{Gaku2019_panopticfusion} & PQ & \textbf{33.5} & \textbf{30.8} & \textbf{58.4}\\
    Ours (NN mapping) & PQ & 31.5 & 30.2 & 43.4\\
    \hdashline
    Ours (skipped missing) & PQ & 36.3 & 34.7 & 51.0\\
    \\[0.1cm]\hline
    PanopticFusion~\cite{Gaku2019_panopticfusion} & SQ & \textbf{73.0} & \textbf{73.3} & 70.7\\ 
    Ours (NN mapping) & SQ & 72.9 & 73.0 & \textbf{72.6}\\
    \hdashline
    Ours (skipped missing) & SQ & 76.1 & 75.9 & 77.9\\
    \\[0.1cm]\hline
    PanopticFusion~\cite{Gaku2019_panopticfusion} & RQ & \textbf{45.3} & \textbf{41.3} & \textbf{80.9}\\
    Ours (NN mapping) & RQ & 42.2 & 40.3 & 59.3\\
    \hdashline
    Ours (skipped missing) & RQ & 46.8 & 44.8 & 64.7\\
    \end{tabular}
    }
    \end{center}
    \caption{3D panoptic segmentation results on the ScanNet v2 open test set. We report the numbers of PanopticFusion~\cite{Gaku2019_panopticfusion} and our fusion method either by a NN mapping or by skipping those missing regions. \textit{Our (NN mapping)} outperforms PanopticFusion in 7 classes, more information can be found in the supplementary material. Note that \textit{Our (skipped missing)} is not considered when highlighting the best score since it‘s not directly comparable.}
    \label{tab:panoptic}
\end{table}%
\subsection{3D Panoptic/Semantic Segmentation}\label{sec:semantic_panoptic_segmentation}%
\definecolor{wall_blue}{RGB}{174,199,232}
\definecolor{floor_green}{RGB}{152,232,138}
\begin{figure*}[t]
    \centering
    \includegraphics[width=\textwidth]{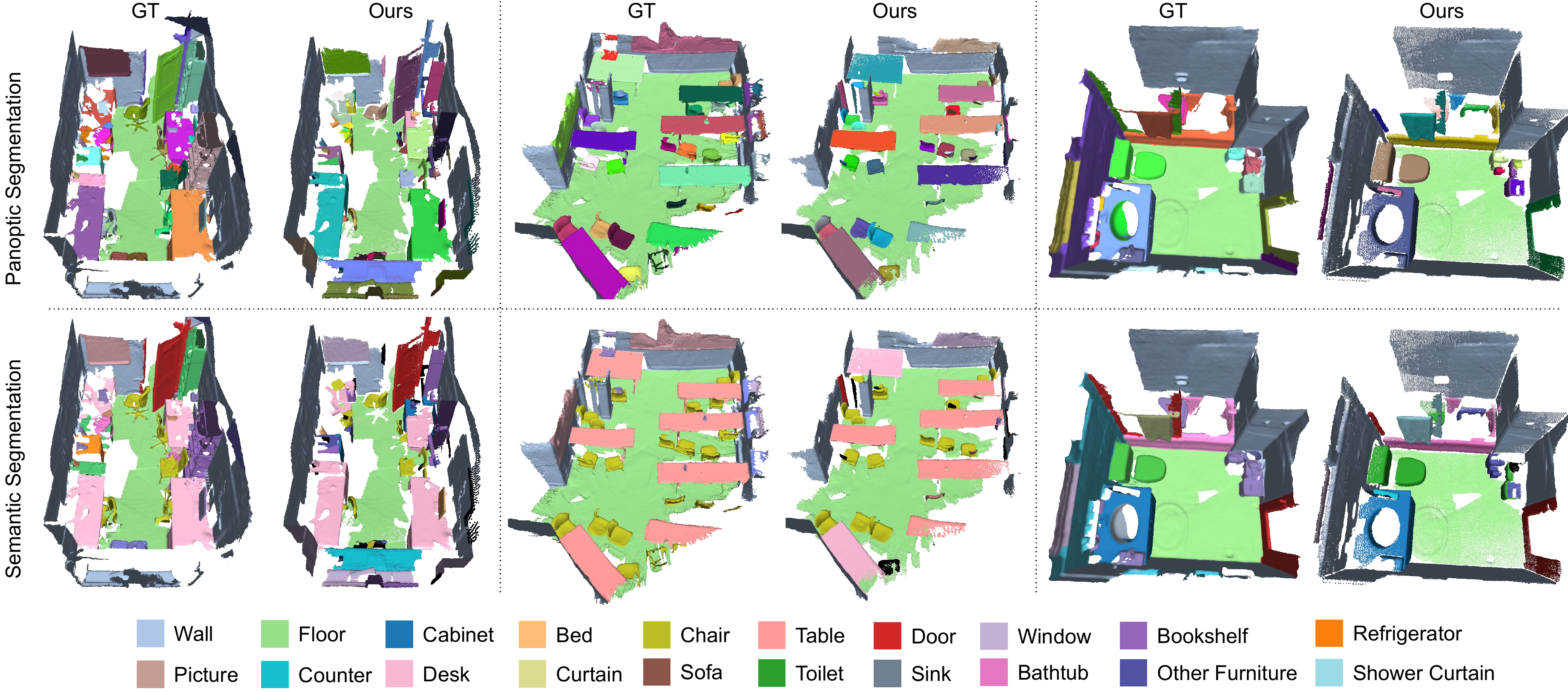}
        \caption{Qualitative semantic and panoptic segmentation results of SceneGraphFusion. Note that for 3D panoptic segmentation (top row), random colors are used for object instances, while walls are \fcolorbox{wall_blue}{wall_blue}{\rule{0pt}{2pt}\rule{2pt}{0pt}} and the floor is \fcolorbox{floor_green}{floor_green}{\rule{0pt}{2pt}\rule{2pt}{0pt}}.}
    \label{fig:semantic_panoptic}
\end{figure*}%
To evaluate the quality of the semantic/panoptic segmentation of our method, we trained the network with ScanNet~\cite{Dai2017_scannet}. We follow the ScanNet benchmark and evaluate with the IoU metric. Since InSeg~\cite{Tateno2015} reconstructs and segments the scene with a different reconstruction algorithm and excludes small and unstable geometric segments, some points might be missing in our 3D map.
When evaluating, we address this issue by either a) mapping the points in our reconstruction to the nearest neighbor (NN) of the ScanNet ground truth 3D model or b) ignoring points where no corresponding 3D geometry was reconstructed.\par%
\paragraph{3D Semantic Segmentation.} In Tbl.~\ref{tab:macc_scannet}, we compare our method against other incremental semantic segmentation methods, specifically SemanticFusion~\cite{McCormac2017_semanticfusion}, ProgressiveFusion~\cite{pham2019real} and FusionAware~\cite{Jiazhao2020_fusionaware} using the mean average precision (mAP). Our method has the second best mAP while running at \methodHz Hz on a CPU, as detailed in Sec.~\ref{sec:runtime} and the supplementary material. Qualitative results of our semantic segmentation are shown in the bottom row of Fig.~\ref{fig:semantic_panoptic}.\par%
\setlength{\tabcolsep}{2pt}
\begin{table}[ht]
    \begin{center}
    \resizebox{\linewidth}{!}{
    \begin{tabular}{p{3.6cm}|p{1.9cm}cP{1cm}}
        & Hardware & Runtime [Hz] & mAP \\
        \hline
SemanticFusion~\cite{McCormac2017_semanticfusion} & GPU + CPU & ~25 & 51.8\\
ProgressiveFusion~\cite{pham2019real}            & GPU + CPU & 10 -- 15 & 56.6\\
Fusionaware~\cite{Jiazhao2020_fusionaware}       & -       & ~10 & \textbf{76.4}\\
Ours                                             & CPU     & ~\textbf{\methodHz} & 63.7\\
    \end{tabular}}
    \end{center}
    \caption{Comparison of incremental semantic segmentation methods on the open test set of ScanNet~\cite{Dai2017_scannet}. Runtime have been taken from the respective papers with potentially different hardware setups, therefore not directly comparable.}
    \label{tab:macc_scannet}
\end{table}%
\paragraph{3D Panoptic Segmentation.} To evaluate panoptic segmentation we use the metrics proposed in~\cite{Kirillov2019}, namely panoptic quality (PQ), segmentation quality (SQ), and recognition quality (RQ). In Tbl.~\ref{tab:panoptic} we compare our method against PanopticFusion~\cite{Gaku2019_panopticfusion}. Due to the missing scene geometry on which our approach relies, PanopticFusion outperforms our method with respect to the computed RQ. Nevertheless, SQ and PQ are on par or slightly worse. A comparison of only valid scene regions -- by skipping un-reconstructed parts -- often results in a better performance. We provide an ablation study in Tbl.~\ref{tab:effectiveness_same_part} to validate the effectiveness of the \texttt{same part} relationship and our proposed fusion mechanism. Finally, the qualitative results of our panoptic segmentation are shown in the top row of Fig.~\ref{fig:semantic_panoptic}.\par%
\setlength{\tabcolsep}{2pt}
\begin{table}[t]
    \begin{center}
    \resizebox{\linewidth}{!}{
    \begin{tabular}{p{5.0cm}|P{1.2cm}P{1.2cm}P{1.2cm}}
         & PQ & SQ & RQ \\
        \hline
        \textcircled{\raisebox{-0.4pt}{\scriptsize 1\normalsize}} Ours without Fusion & 35.4 & 76.3 & 35.4\\
        \textcircled{\raisebox{-0.4pt}{\scriptsize 2\normalsize}} Ours without \texttt{same part} & 10.9 & 59.1 & 16.0 \\
        \textcircled{\raisebox{-0.4pt}{\scriptsize 3\normalsize}} Ours & 36.3 & 76.1 & 46.8 \\
    \end{tabular}}
    \end{center}
    \caption{Ablation study: Analysis of the effect of our fusion mechanism and the \texttt{same part} relationship on the panoptic segmentation task evaluated on ScanNet~\cite{Dai2017_scannet}. PQ stands for panoptic, SQ for segmentation, and RQ for recognition quality.}
    \label{tab:effectiveness_same_part}
\end{table}%
\paragraph{Robustness against Missing Information.}\label{sec:drop_edge}%
In this experiment, we evaluate the robustness of different attention methods against noisy data in form of missing edges. We train our network without attention, using GAT~\cite{Velickovi2018_GAT}, SDPA~\cite{Ashish2017_attention}, and our proposed FAT. The experimental setup is shared with Tbl.~\ref{tab:3dssg_inseg_20_8}. In Tbl.~\ref{tab:eva_dropedge} we compare the performance of the different attention methods on the full scene (\(f\)) and all edges (@1.0, left column) and with a random edge drop of 50\% (@0.5, right column), see Tbl.~\ref{tab:eva_dropedge}.\par%
The reference metric is the intersection over union (IoU). Our proposed attention mechanism, FAT, consistently outperforms the other approaches in full-edge and drop-edge scenarios. For the sake of space, a more detailed per-class evaluation is available in the supplementary, interestingly showing that some classes rely on the messages from neighbors more than others such as \eg bathtub, shower curtain, and windows.\par%
\setlength{\tabcolsep}{2pt}
\begin{table}[t]
\begin{center}
    \small
    \begin{tabular}{p{2.5cm}|P{2.9cm}P{2.8cm}}
    & avg. IoU (@1.0) & avg. IoU (@0.5)\\
   \hline
Ours w/o attention & 33.5 & 29.5 \\ 
Ours SDPA~\cite{Ashish2017_attention} & 33.0 & 29.7\\ 
Ours GAT~\cite{Velickovi2018_GAT} &  11.5 & 12.5\\
Ours FAT & \textbf{49.3}  & \textbf{41.9}\\
    \end{tabular}
    \end{center}
    \caption{Ablation study: Segment classification of InSeg~\cite{Tateno2015} on 3RScan~\cite{Wald2019RIO} reporting avg. IoU on segment-level. The complete per-class evaluation can be found in the supplementary material.}
    \label{tab:eva_dropedge}
\end{table}%
\setlength{\tabcolsep}{2pt}
\begin{table}[t]
    \begin{center}
    \begin{tabular}{p{1.7cm}|P{1.6cm}P{1.4cm}P{1.4cm}P{1.4cm}}
        & Segmentation & Node & Edge & GNN \\
      \hline
      Mean [ms] & 28 & 8 & 17 & 108 \\
    \end{tabular}
   \end{center}
    \caption{Runtime [ms] of the different components of our method.}
    \label{tab:time}
\end{table}%
\subsection{Runtime Analysis}\label{sec:runtime}%
We measured the runtime of our system on the ScanNet sequence \texttt{scene0645\_01}. Our machine is equipped with an Intel Core i7-8700 CPU 3.2GHz CPU with 12 threads. Notably, our method only uses 2 threads: one for the scene reconstruction and the other one for 3D scene graph prediction. The scene reconstruction requires $28$ ms on average while the graph prediction sums up to $133$ms running the GNN and fusing the results.\par%
\section{Conclusion}%
In this work, we presented SceneGraphFusion, a 3D scene graph method that incrementally fuses partial graph predictions from a geometric segmentation into a globally consistent semantic map. Our network outperforms other 3D scene graph prediction methods; FAT works better than any other attention mechanism in handling missing graph information and the semantic/panoptic segmentation -- the by-product of our method -- achieves performance on par with other incremental methods while running in \methodHz Hz. Due to this efficiency, incremental semantic scene graphs could be beneficial in future work, when retrieving camera poses or detecting loop-closures in a SLAM framework.\par%
\section*{Acknowledgment}\label{sec:Acknowledgment}%
The authors thank Anees B. Kazi and Georam Vivar for fruitfull discussions. This work is supported by the \emph{German Research Foundation} (DFG, project number 407378162) and the Bavarian State Ministry of Education, Science and the Arts in the framework of the Centre Digitisation Bavaria (ZD.B).\par%
\newpage
\section{Supplementary Material}
\subsection{Network Architecture}
We use \(\text{FC}(\textit{in}, \textit{out})\) denote a fully-connected layer, and \(\text{MLP}(\cdot, ..., \cdot)\) as a set of FC layers with a ReLU activation between each FC layer. Our PointNet encoder \(f_p\), is a shared-weight \(\text{MLP}(64,128,512)\) followed by a maximum pooling operation to obtain a global feature. The other layers are listed in Tbl.~\ref{tab:network_params}.\par
\subsection{Training Details}
All the models in our evaluation section (Sec.~\ref{sec:evaluation}) were trained with the same set up but with different training data for $150$ epochs. We use AdamW \cite{Loshchilov2017decoupled} optimizer with Amsgrad \cite{Reddi2018convergence} and an adaptive learning rate, inverse proportional to the log of the number of edges. Given a training batch with $n$ edges and \(\text{lr}_{base} = 1e^{-3}\), the base learning rate in AdamW is adjusted as follows
\begin{equation}
    \text{lr} = \text{lr}_{base} \frac{1}{\ln{n}}.
\end{equation}
The training data are the 3D reconstructions created from RGB-D sequences. In order to train our network to handle partial data, subgraphs are randomly extracted during training time. In each iteration, two segments are randomly selected together with their four-hop neighbor segments. We further randomly discard edges with a dropout rate of $50\%$. In addition, we randomly sample points in each segment. The properties described in Sec~\ref{sec:inseg} are computed based on sampled points. For the training loss, we follow the approach in \cite{Wald2020_3dssg} with a weight factor of $0.1$ between the object and predicate loss. We use two message passing layers, each with 8 heads.\par
\subsection{Experiment Details}
In this section, we detailed the training dataset and the hyper-parameters used in the experiment section (Sec~\ref{sec:eval_scene_graph_pred} Geometric Segments) of our main paper. As mentioned in the main paper, 20 NYUv2~\cite{Silberman2012} object classes are used. For predicates, we focus on \textit{support} relationships. We further filter out rare relationship. A predicate is discarded if it occurs less than 10 times in the training data or less then 5 times in the test data. This leaves us with 8 predicates, \ie supported by, attached to, standing on, hanging on, connected to, part of, build in, and same part.\par
The geometrical segmentation method~\cite{Tateno2015} in our framework 
uses the pyramid level of 2, which scales the input image with a factor of 2, for image segmentation. Further, we filter out segments with less than 512 points.\par
%
\subsection{3D Panoptic Segmentation}
On Tbl.~\ref{tab:panoptic_full} we report the complete panoptic segmentation evaluation on Tbl.~\ref{tab:panoptic}. With respect to the panoptic quality (PQ), our method outperforms PanopticFusion in 7 out of 20 classes. The PQ can be broken down into segmentation quality (SQ) and recognition quality (RQ). The SQ evaluates only the matched segments, via an intersection over union (IoU) score over $50\%$. RQ is known as the \(F_1\) score. Our method has a similar SQ performance as PanopticFusion while performing worse when compared with the RQ metric. This is likely due to missing scene geometry caused by the incremental segmentation~\cite{Tateno2015} that our approach relies on. By using a metric that is less influenced by missing points, \ie SQ, or ignoring the missing points in the evaluation, our method has equivalent or slightly better performance compare to PanopticFusion~\cite{Gaku2019_panopticfusion}.\par
%
\subsection{Robustness against Missing Information}
Tbl.~\ref{tab:eva_dropedge_robust} shows the complete experiment mentioned in Sec~\ref{sec:semantic_panoptic_segmentation} of the main paper. We use our network architecture with different attention methods. The update of the node feature \(v_i\) in equation~7 can be re-written as follows:%
\begin{equation} \label{eq:abla_attention}
    \mathbf{v}^{\ell+1}_{i} = g_{e}\left( [\mathbf{v}^{\ell}_{i},
    \Phi_{j\in \mathcal{N}\left(i\right)}\left( \Psi\left( \cdot \right) \right) ]\right),
\end{equation}%
where \(\Phi\) is a permutation in-variance function, \eg sum, mean or max, and \(\Psi(\cdot)\) represents an attention function. 
For \textit{without}, we set \(\Psi(\cdot)\) to \(\Psi(\mathbf{v}^{\ell}_{j})=\mathbf{v}^{\ell}_{j}\) with \(\Phi=\sum\). For \textit{SDPA}, the \(\Psi(\cdot)\) is set to \( f_{sdpa}(\mathbf{v}^{\ell}_{i}, \mathbf{v}^{\ell}_{j}) \) with \(\Phi=\sum\), where \(f_{sdps}\) is the multi-head attention method~\cite{Ashish2017_attention} and for \textit{GAT}, we directly use the Pytorch implementation~\cite{pytorch_geometric} to update nodes respectively.\par
\begin{figure}
    \centering
    \includegraphics[width=\columnwidth]{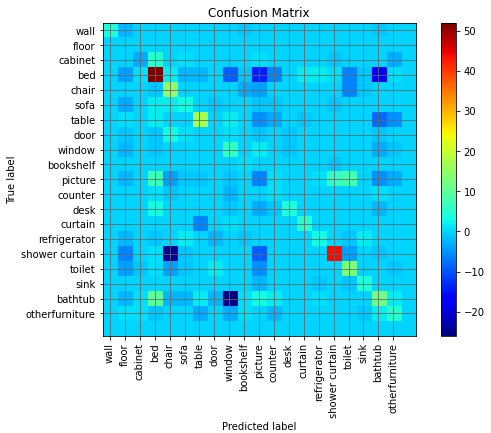}
    \caption{The difference of the confusion matrix without and with (50\%) dropping of edges.}
    \label{fig:confusion_dropedge}
\end{figure}

The proposed attention method consistently outperforms others even when the edges are dropped by 50\%. There are some classes that are more robust to missing information, such as a chair, curtain, desk, floor, and wall while some classes are more dependant on others such as bathtub, bed, shower curtain, and window. We visualize this effect by plotting the difference of the confusion matrix with and without dropping of edges, see Fig.~\ref{fig:confusion_dropedge}. It can be seen that beds and pictures are easier to predict when full edges are provided.\par
%
\subsection{Runtime Analysis}
In the following a more detailed runtime analysis is given in terms of the number of segments and data re-usage and graph structure update.\par
\begin{figure*}
    \begin{center}
    \includegraphics[width=0.75\textwidth]{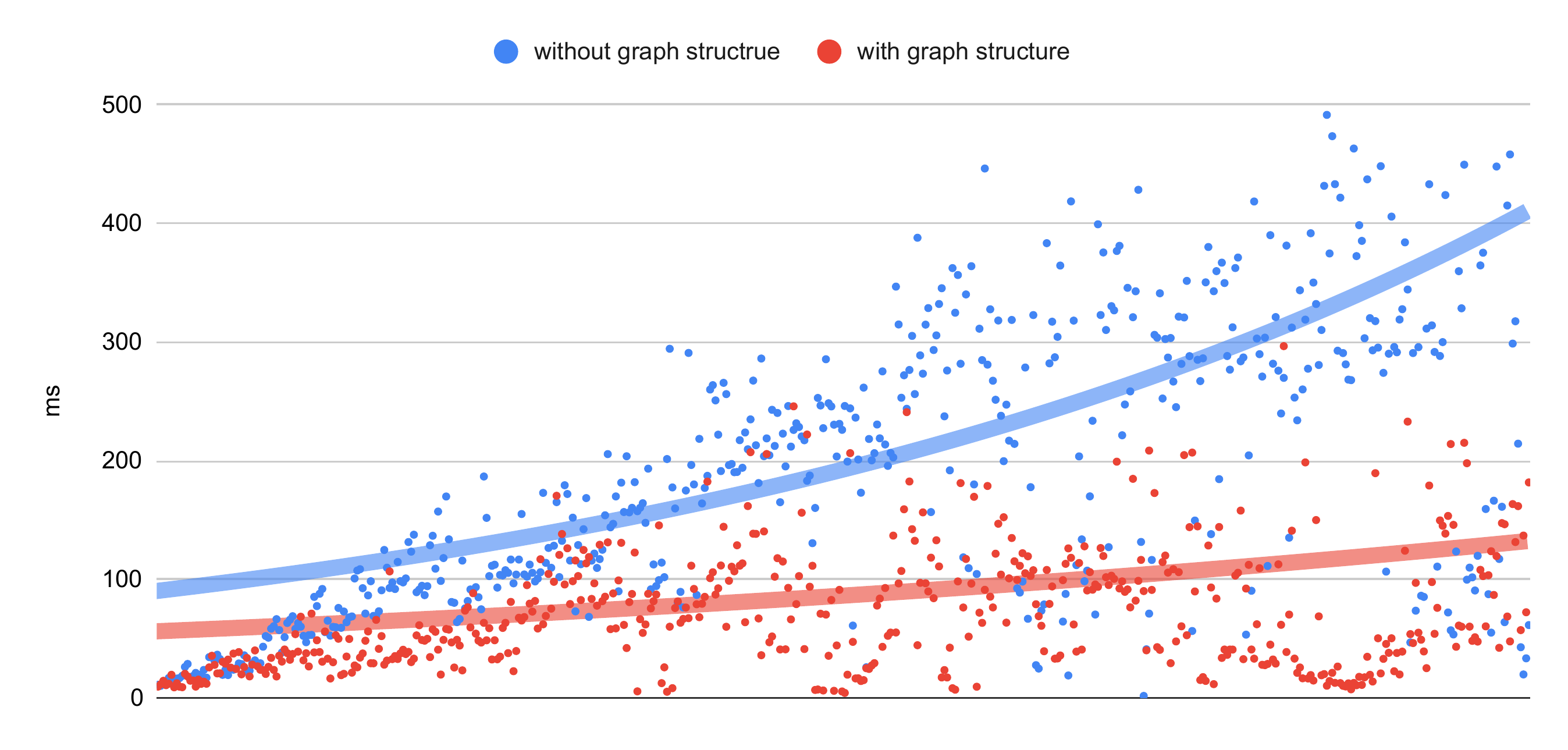}
    \end{center}
    \caption{The computation time of the scene graph prediction over time.
    }
    \label{fig:avg_time_overtime}
\end{figure*}

We report the analysis using scene \texttt{scene0645\_01} which consists of 5230 paired RGB and depth images. The average update of node, edge, GNN features and the class predictions are listed on Tbl.~\ref{tab:time_graph_update}. The computation time over-time is reported on Fig.~\ref{fig:avg_time_overtime}. By updating node and edge features with our graph structure, the computation time is significantly reduced. Again, our scene graph prediction method runs in a different thread and will only block the main thread in the data copy and fusion stage.\par
\begin{table}[t]
    \begin{center}
    \begin{tabular}{l|rr}
         & \# computations & times (ms) \\
         \hline
        Node Feature & 2.13 & 2.49 \\
        Edge Feature & 20.83 & 1.53 \\
        GNN Feature 1 & 20.83 & 14.28 \\
        GNN Feature 2 & 57.88 & 44.06\\
        Class Prediction & 57.88 & 9.74 \\
    \end{tabular}
    \end{center}
    \caption{The average number and time of computation on each feature computation process on the sequence of \texttt{scene0645\_01}}
    \label{tab:time_graph_update}
\end{table}
%
\subsection{Qualitative Result}
We demonstrate more qualitative results in the 3D scene graph prediction on both 3RScan~\cite{Wald2019RIO} and ScanNet~\cite{Dai2017_scannet}. Note that ScanNet does not have ground truth relationships. We therefore use the trained model with 3RScan to do inference on ScanNet scenes. Our method is able to handle the domain gap across these two datasets and predicts reasonable 3D scene graphs on ScanNet scenes.\par
\begin{figure*}
    \centering
    \includegraphics[width=0.98\textwidth]{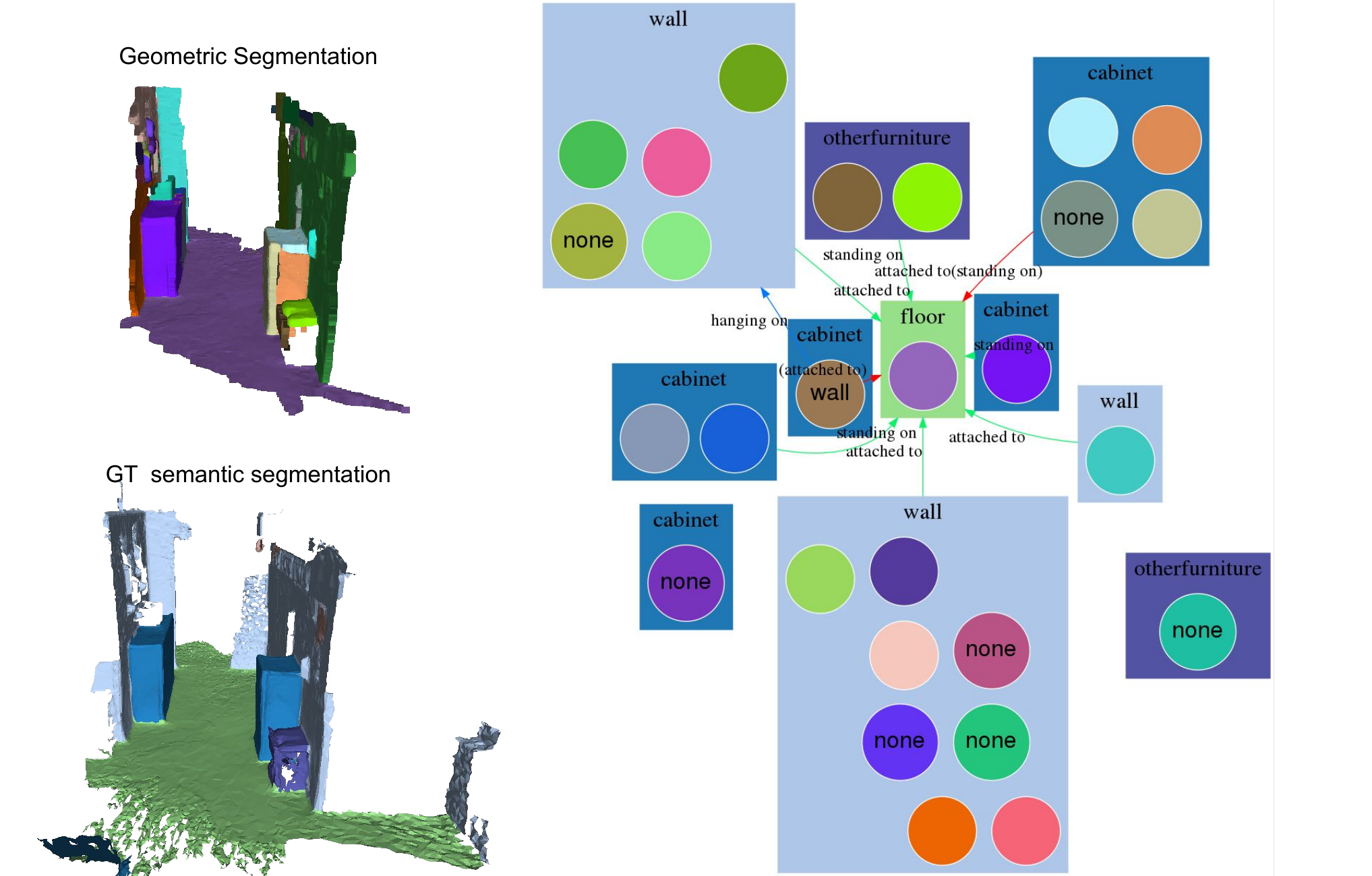}
    \includegraphics[width=0.98\textwidth]{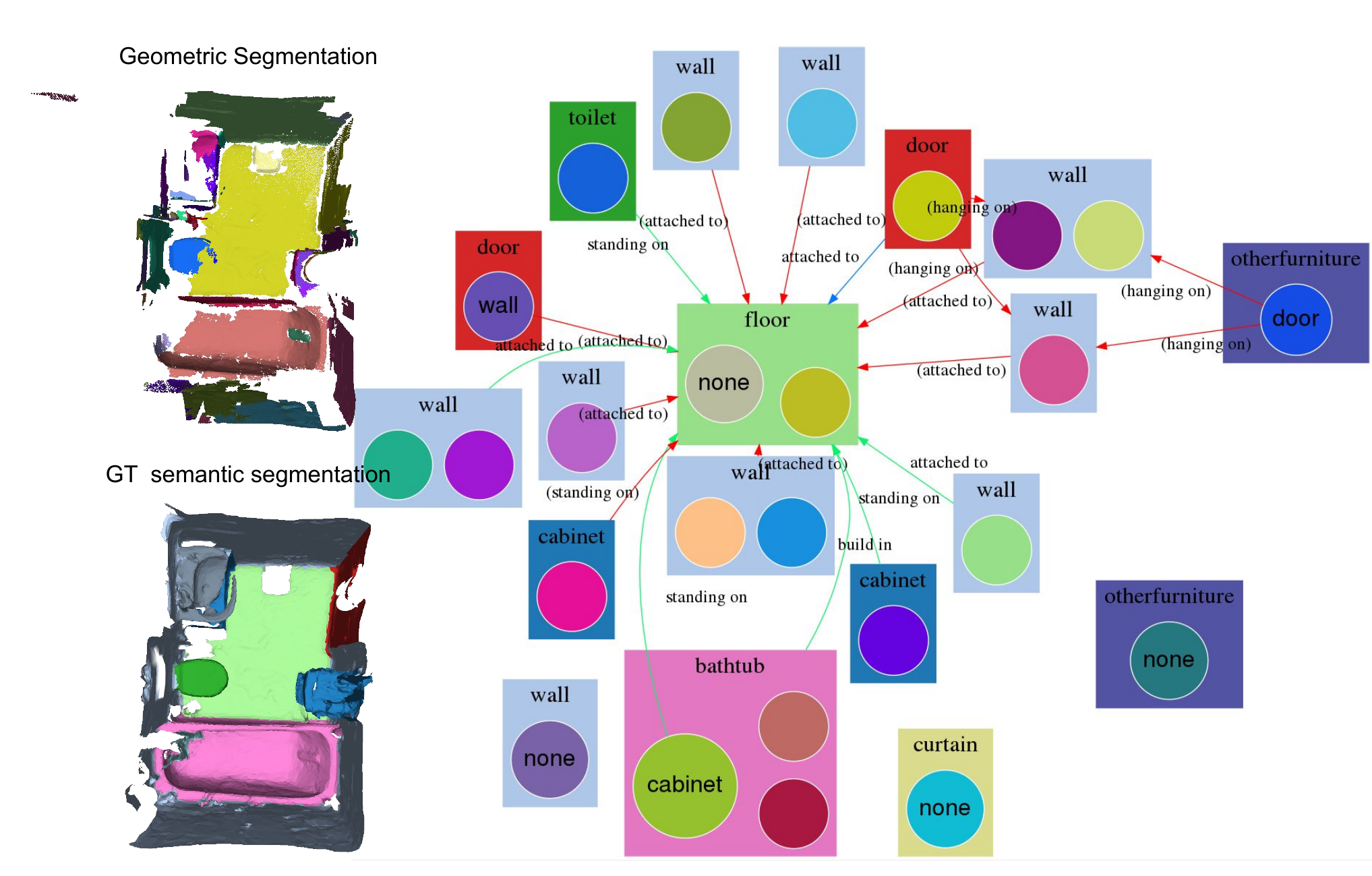}
    \caption{Qualitative results of our method on example scenes from 3RScan~\cite{Wald2019RIO}.}
    \label{fig:qualitative_3rscan_small}
\end{figure*}
\begin{figure*}
    \centering
    \includegraphics[width=0.98\textwidth]{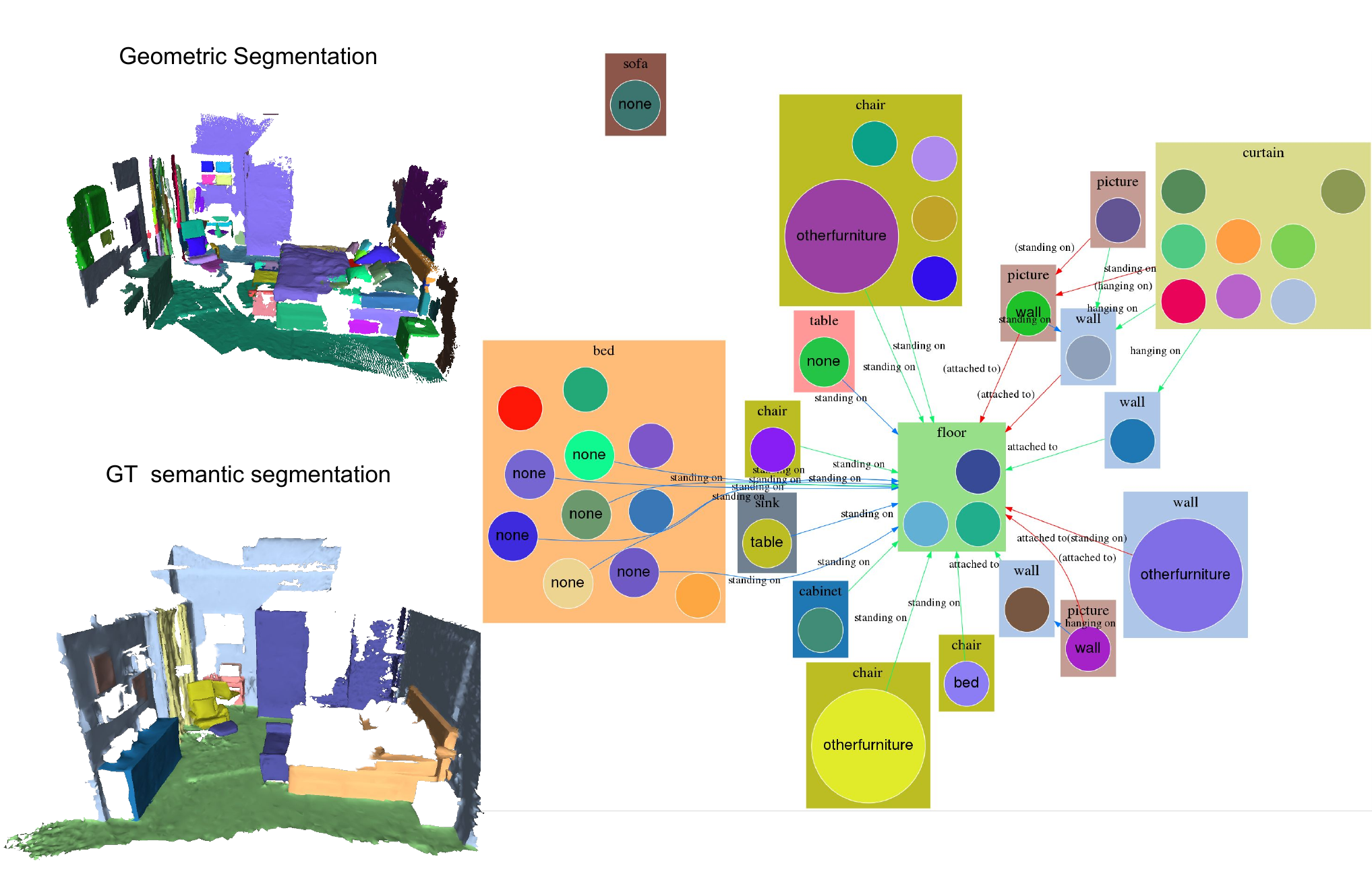}
    \includegraphics[width=0.98\textwidth]{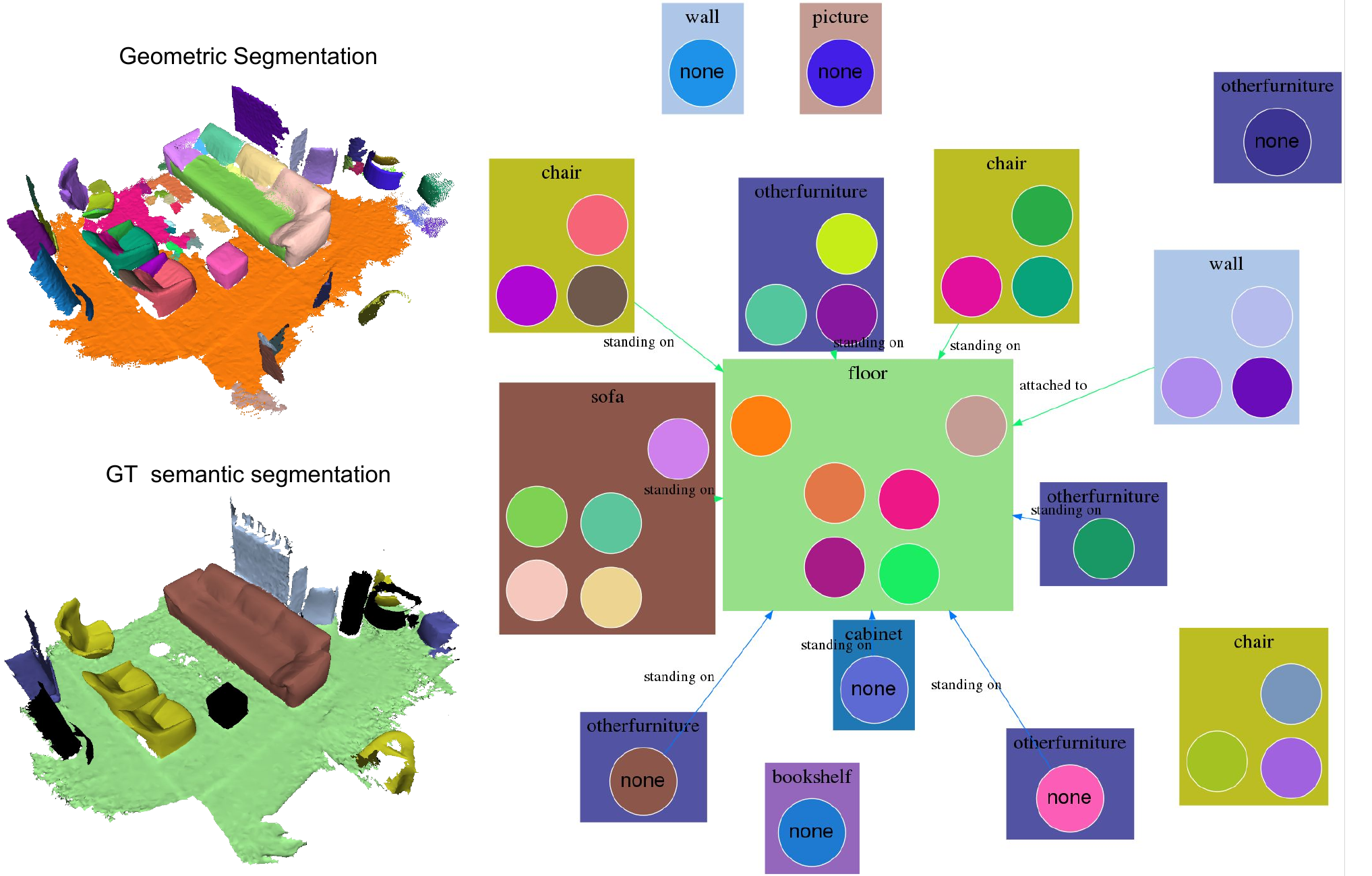}
    \caption{Qualitative results of our method on relative large scenes from 3RScan~\cite{Wald2019RIO}.}
    \label{fig:qualitative_3rscan_larger}
\end{figure*}
\begin{figure*}
    \centering
    \includegraphics[width=0.98\textwidth]{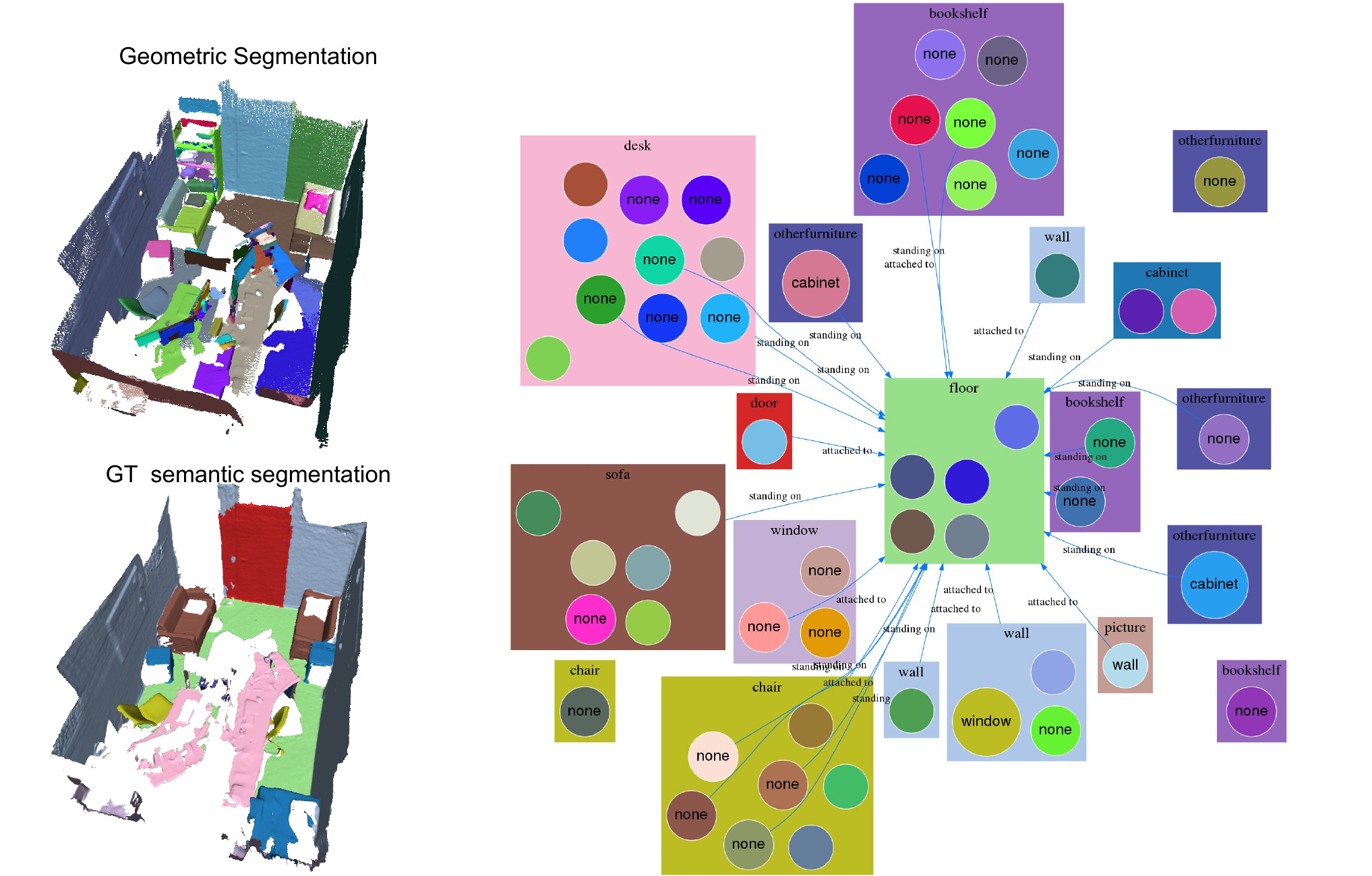}
    \includegraphics[width=0.98\textwidth]{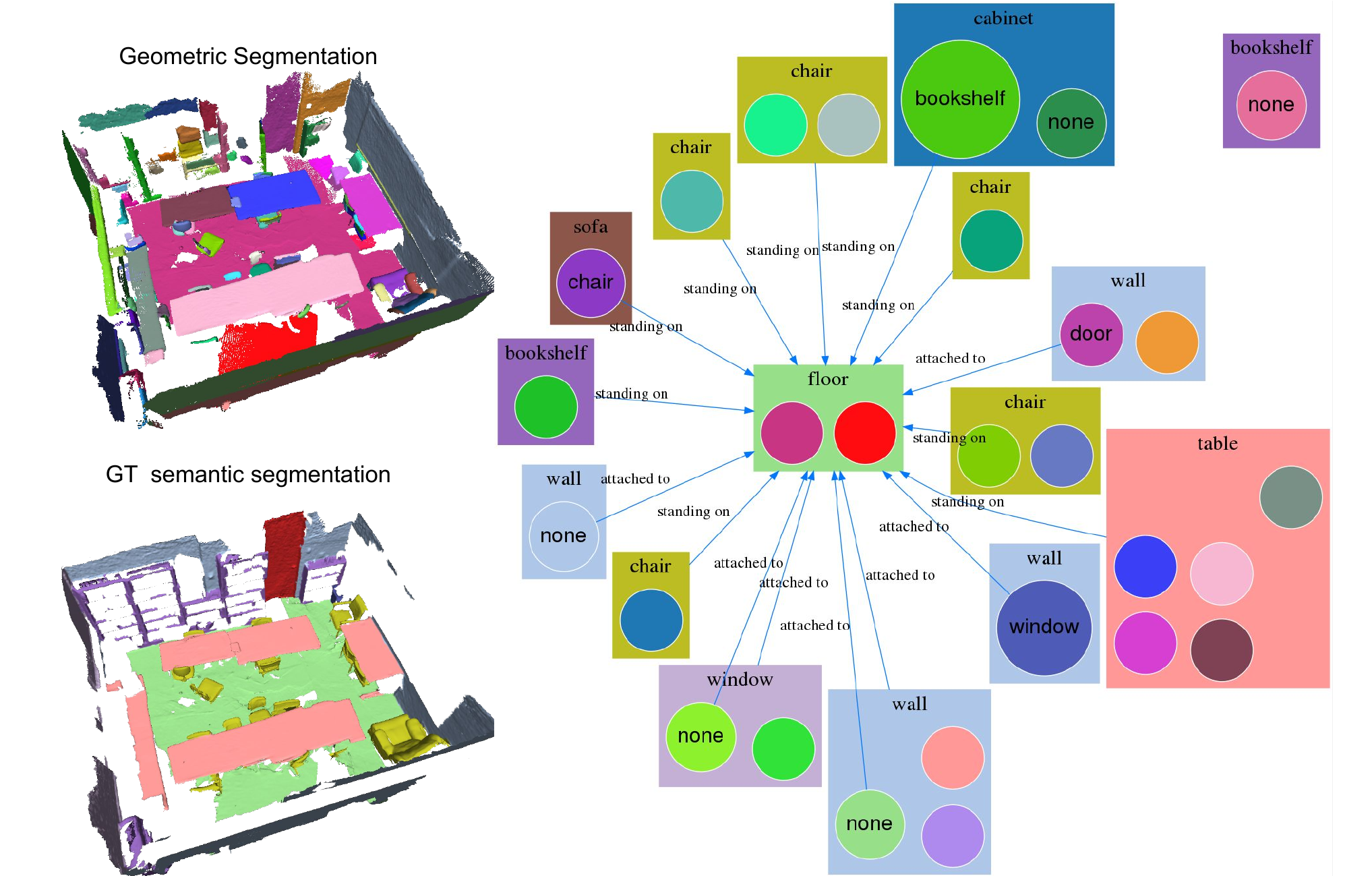}
    \caption{Qualitative results of our method on the scenes from ScanNet~\cite{Dai2017_scannet}.}
    \label{fig:qualitative_scannet}
\end{figure*}

The results are shown on Fig.~\ref{fig:qualitative_3rscan_small}, Fig.~\ref{fig:qualitative_3rscan_larger} and Fig.~\ref{fig:qualitative_scannet}. Segments are represented by circles and estimated object instances are drawn as rectangles. In our visualization, the class prediction of a segment is correct if no label is shown in the circle and wrong otherwise.\par
As for relationship prediction, we use green, red and blue to indicate the correct, wrong, and unknown predictions respectively. An unknown prediction is a case where no ground truth data is available. The label on an edge without bracket is the predicted label, with bracket is its ground truth label. To simplify the visualization, we ignore \textit{none}-relationships and merge segments with \textit{same part} relationships in the same box. We also group up predictions of segments with the same label within the same box. The indication of such a grouped prediction is shown by connecting box to box. As for the wrongly predicted segments, their predicted probability remains individual. This indicated with an edge from a circle to a box.\par
%
\begin{table}[th!]
    \begin{center}
    \begin{tabular}{l|l}
        Function & Layer Definition \\
        \hline
         \(g_{v}, g_{a}\) & MLP(768, 768, 512) \\
         \(g_{e}\) & MLP(1280, 768, 256) \\
         \(\hat{g}_{q}, \hat{g}_{e}\) & FC(512, 512)\\
         \(\hat{g}_{\tau}\) & FC(256, 256)\\
    \end{tabular}
    \end{center}
    \caption{Parameters of the layers in our GNN. FC\((\cdot,\cdot)\) represents fully connected layer, and MLP\((\cdot,...,\cdot)\) represent FC\((\cdot,\cdot)\) layers with ReLU activation between them.}
    \label{tab:network_params}
\end{table}
\newcolumntype{R}[1]{>{\RaggedLeft\arraybackslash}p{#1}}
\setlength{\tabcolsep}{0.8pt}
\begin{table*}[!htbp]
    \begin{center}
    \small
    \begin{tabular}{p{2.0cm}R{0.65cm}R{0.65cm}R{0.65cm}R{0.65cm}R{0.65cm}R{0.65cm}R{0.65cm}R{0.65cm}R{0.65cm}R{0.65cm}R{0.65cm}R{0.65cm}R{0.65cm}R{0.65cm}R{0.65cm}R{0.65cm}R{0.65cm}R{0.65cm}R{0.65cm}R{0.65cm}R{0.65cm}}
    & bath & bed & bkshf & cab. & chair & cntr. & curt. & desk & door & floor & ofurn & pic. & refri. & show. & sink & sofa & table & toil & wall & wind. & avg \\
        \hline
        without       & 50.0 & 3.9 & 0.0 & 27.0 & 51.7 & 16.7 & 62.2 & \textbf{20.0} & 16.4 & 96.2 & 15.4 & 8.0 & 4.3 & 11.1 & 52.5 & 45.9 & \textbf{54.2} & 41.7 & 67.0 & 26.2 & 33.5\\
SDPA\cite{Ashish2017_attention}      & 50.0 & 11.1 & \textbf{2.3} & 26.0 & 45.7 & 17.7 & 65.2 & 3.9 & 18.7 & 87.4 & 11.2 & 4.8 & 2.5 & 29.4 & 38.1 & 60.8 & 36.8 & 65.0 & 60.1 & 24.0 & 33.0\\
GAT\cite{Velickovi2018_GAT} & 22.0 & 5.7 & 0.0 & 10.9 & 22.8 & 11.0 & 37.6 & 1.8 & 9.4 & 19.8 & 3.1 & 1.3 & 0.0 & 0.0 & 10.4 & 33.0 & 12.0 & 8.7 & 8.7 & 11.9 & 11.5\\
        ours      & \textbf{83.3} & \textbf{24.3} & 0.0 & \textbf{43.4} & \textbf{69.8} & \textbf{30.0} & \textbf{68.7} & 4.5 & \textbf{29.6} & \textbf{98.1} & \textbf{26.6} & \textbf{10.0} & \textbf{34.5} & \textbf{66.7} & \textbf{65.0} & \textbf{74.7} & \textbf{54.2} & \textbf{86.5} & \textbf{75.3} & \textbf{41.7} & \textbf{49.3}\\
        \hline
        without@p50  & 37.5 & 3.7 & 0.0 & 24.8 & 49.9 & 13.3 & 52.3 & \textbf{17.3} & 14.9 & 86.8 & 19.9 & 5.0 & 5.8 & 6.2 & 46.7 & 36.4 & 37.3 & 46.0 & 63.5 & 22.8 & 29.5\\
        SDPA@p50 & 45.5 & 8.5 & 0.0 & 24.2 & 45.0 & 8.8 & \textbf{59.9} & 6.5 & 16.1 & 81.2 & 11.0 & 4.0 & 3.4 & 23.1 & 35.5 & 54.2 & 36.3 & 47.7 & 59.6 & 23.1 & 29.7\\
GAT\cite{Velickovi2018_GAT}@p50 & 26.5 & 2.3 & 0.0 & 14.6 & 21.9 & 4.6 & 34.9 & 0.0 & 8.4 & 18.6 & 4.7 & 1.2 & 7.1 & 17.6 & 7.9 & 30.2 & 11.2 & 8.9 & 17.1 & 12.9 & 12.5\\
        ours@p50 & \textbf{61.1} & \textbf{14.3} & \textbf{7.9} & \textbf{35.1} & \textbf{62.0} & \textbf{23.8} & 59.5 & 4.3 & \textbf{23.4} & \textbf{96.7} & \textbf{25.6} & \textbf{6.7} & \textbf{19.4} & \textbf{41.2} & \textbf{63.2} & \textbf{65.2} & \textbf{47.4} & \textbf{73.7} & \textbf{72.2} & \textbf{34.6} & \textbf{41.9}\\
    \end{tabular}
    \end{center}
    \caption{Ablation study: Segment classification of InSeg~\cite{Tateno2015} on 3RScan~ \cite{Wald2019RIO} reporting avg.  IoU on segment-level.}
    \label{tab:eva_dropedge_robust}
\end{table*}%
\setlength{\tabcolsep}{0.8pt}
\begin{table*}[!ht]
    \small
    \begin{center}
    \resizebox{\textwidth}{!}{
    \begin{tabular}{lc|ccc|R{0.65cm}R{0.65cm}R{0.65cm}R{0.65cm}R{0.65cm}R{0.65cm}R{0.65cm}R{0.65cm}R{0.65cm}R{0.65cm}R{0.65cm}R{0.65cm}R{0.65cm}R{0.65cm}R{0.65cm}R{0.65cm}R{0.65cm}R{0.65cm}R{0.65cm}R{0.65cm}R{0.65cm}}
            & metric & all & things & stuff & bath & bed & bkshf & cab. & chair & cntr. & curt. & desk & door & floor & ofurn & pic. & refri. & show. & sink & sofa & table & toil & wall & wind.\\
            \hline
    PanopticFusion~\cite{Gaku2019_panopticfusion} & PQ     & 33.5     & \textbf{30.8}     & \textbf{58.4}     & 31.0         & \textbf{35.8}           & \textbf{16.4}        & \textbf{23.8}      & 46.7       & 10.4         & \textbf{16.6}      & \textbf{16.1}            & 18.0      & \textbf{76.4}     & \textbf{27.7}       & 26.4         & \textbf{39.5}     & 36.3       & 36.7      & \textbf{42.1}            & \textbf{34.8}     & \textbf{76.1}     & \textbf{40.4}            & \textbf{19.3}     \\
    Ours (NN mapping)    & PQ     & 31.5     & 30.2        & 43.4       & \textbf{67.6}      & 25.4     & 13.9           & 22.2         & \textbf{47.2}     & \textbf{10.5}      & 16.4         & 12.6      & \textbf{26.4}            & 56.4       & 22.9       & \textbf{31.3}      & 28.0     & \textbf{38.3}       & \textbf{38.0}            & 32.3      & 34.8       & 63.2        & 30.4      & 11.7        \\
    Ours (skip missing)    & PQ     & 36.3     & 51.0        & 34.7       & 68.4         & 28.0     & 16.0           & 26.4         & 58.1       & 15.6         & 24.7         & 17.7      & 28.7      & 64.5       & 26.9       & 35.4         & 30.8     & 40.7       & 41.3      & 38.8      & 45.6       & 66.2        & 37.4      & 15.2        \\
    \hline
    PanopticFusion~\cite{Gaku2019_panopticfusion} & SQ     & \textbf{73.0}           & \textbf{73.3}     & 70.7       & 75.3         & \textbf{70.1}           & \textbf{73.9}        & \textbf{71.1}      & 74.3       & \textbf{65.1}      & 72.3         & 61.7      & 76.0      & \textbf{77.4}     & \textbf{75.8}       & 71.2         & 77.7     & \textbf{79.5}       & 72.7      & \textbf{74.6}            & \textbf{74.3}     & \textbf{81.4}     & 64.0      & \textbf{72.5}     \\
    Ours (NN mapping)    & SQ     & 72.9     & 73.0        & \textbf{72.6}     & \textbf{80.6}      & 68.2     & 66.9           & \textbf{71.1}      & \textbf{76.5}     & 61.7         & \textbf{75.1}      & \textbf{63.8}            & \textbf{77.4}            & 74.8       & 71.6       & \textbf{81.5}      & \textbf{77.8}     & 79.1       & \textbf{75.4}            & 65.3      & 73.3       & 80.2        & \textbf{70.4}            & 68.2        \\
    Ours (skip missing)    & SQ     & 76.1     & 77.9        & 75.9       & 82.9         & 71.2     & 69.1           & 74.6         & 81.2       & 62.3         & 74.0         & 68.0      & 81.0      & 81.8       & 74.4       & 82.7         & 82.3     & 81.5       & 77.2      & 70.2      & 80.9       & 82.4        & 74.0      & 69.5        \\
    \hline
    PanopticFusion~\cite{Gaku2019_panopticfusion} & RQ     & \textbf{45.3}           & \textbf{41.3}     & \textbf{80.9}     & 41.2         & \textbf{51.1}           & \textbf{22.2}        & \textbf{33.5}      & \textbf{62.8}     & 16.0         & \textbf{23.0}      & \textbf{26.0}            & 23.6      & \textbf{98.7}     & \textbf{36.5}       & 37.1         & \textbf{50.8}     & 45.7       & \textbf{50.5}            & \textbf{56.3}            & 46.9       & \textbf{93.5}     & \textbf{63.1}            & \textbf{26.7}     \\
    Ours (NN mapping)    & RQ     & 42.2     & 40.3        & 59.3       & \textbf{83.9}      & 37.2     & 20.7           & 31.3         & 61.7       & \textbf{17.1}      & 21.8         & 19.7      & \textbf{34.1}            & 75.4       & 31.9       & \textbf{38.5}      & 36.0     & \textbf{48.5}       & 50.3      & 49.5      & \textbf{47.5}     & 78.9        & 43.2      & 17.1        \\
    Ours (skip missing)    & RQ     & 46.8     & 64.7        & 44.8       & 82.5         & 39.4     & 23.2           & 35.4         & 71.6       & 25.0         & 33.3         & 26.0      & 35.4      & 78.8       & 36.1       & 42.8         & 37.4     & 50.0       & 53.5      & 55.3      & 56.4       & 80.4        & 50.6      & 21.9       
    \end{tabular}
    }
    \end{center}
    \caption{The full 3D panoptic segmentation results on the ScanNet v2 open test set.}
        \label{tab:panoptic_full}
\end{table*}%
{\small
\bibliographystyle{ieee_fullname}
\bibliography{ms}
}%
\newpage

\end{document}